\documentclass[10pt,twocolumn,letterpaper]{article}

\usepackage[pagenumbers]{iccv} 

\definecolor{iccvblue}{rgb}{0.21,0.49,0.74}
\usepackage[pagebackref,breaklinks,colorlinks,allcolors=iccvblue]{hyperref}

\def\paperID{*****} 
\def\confName{ICCV}
\def\confYear{2025}

\title{Pulling Back the Curtain:\\
       Unsupervised Adversarial Detection via Contrastive Auxiliary Networks}

\author{Eylon Mizrahi\\
Ben-Gurion University\\
Beer-Sheva, 8410501, Israel\\
{\tt\small eylonmiz@post.bgu.ac.il}
\and
Raz Lapid\\
DeepKeep \\
Tel-Aviv, Israel\\
{\tt\small raz.lapid@deepkeep.ai}
\and
Moshe Sipper\\
Ben-Gurion University\\
Beer-Sheva, 8410501, Israel\\
{\tt\small sipper@bgu.ac.il}
}

\usepackage{graphicx} 
\usepackage{ragged2e}
\usepackage{caption}
\usepackage{multirow}
\usepackage{tcolorbox}
\usepackage{amsmath}
\usepackage{amsfonts} 
\usepackage{mdframed}
\usepackage{times}
\usepackage{diagbox}
\usepackage{ragged2e}
\usepackage{booktabs}
\usepackage{tikz}
\usetikzlibrary{shapes,arrows,calc,positioning,decorations.pathmorphing}
\usetikzlibrary{shapes.geometric}

\usepackage{algpseudocode}   
\usepackage[ruled,vlined]{algorithm2e} 

\usepackage{amsmath}         
\usepackage{amssymb}         
\usepackage{amsfonts}        
\usepackage[T1]{fontenc}

\usepackage[numbers,sort&compress]{natbib}
\usepackage{pifont}
\usepackage{float}
\usepackage[font={small}]{caption}
\usepackage[export]{adjustbox}
\usepackage{xcolor,colortbl}
\usepackage{siunitx}
\usepackage{tcolorbox}

\newcommand{\norm}[1]{\left\lVert#1\right\rVert}
\newcommand{\normalize}[1]{\frac{#1}{\norm{#1}}}

\usepackage{sidecap}

\newcolumntype{C}[1]{>{\centering\arraybackslash}m{#1}}
\newcolumntype{G}[1]{>{\columncolor{Gray!30}\centering\arraybackslash}m{#1}}
\newcolumntype{A}[1]{>{\raggedright\arraybackslash}m{#1}}

\begin{document}
\maketitle

\begin{abstract}
Deep learning models are widely employed in safety-critical applications yet remain susceptible to adversarial attacks---imperceptible perturbations that can significantly degrade model performance and challenge detection mechanisms. In this work, we propose an \textbf{U}nsupervised adversarial detection via \textbf{C}ontrastive \textbf{A}uxiliary \textbf{N}etworks (\textbf{U-CAN}) to effectively uncover adversarial behavior within refined feature representations, without the need for adversarial examples. Tested on CIFAR-10, Mammals, and an ImageNet subset with ResNet-50, VGG-16, and ViT backbones, U-CAN outperforms prior unsupervised detectors, delivering the highest average F1 scores against three powerful adversarial attacks. We also evaluate U-CAN under adaptive attacks crafted with full knowledge of both the classifier and detector. The proposed framework provides a scalable and effective solution for enhancing the security and reliability of deep learning systems.
\end{abstract}

\section{Introduction}
\label{sec:introduction}
Deep learning models achieve state-of-the-art performance in domains such as autonomous driving \citep{grigorescu2020survey,bachute2021autonomous}, medical imaging \citep{esteva2019guide,suzuki2017overview}, and cybersecurity \citep{ford2014applications,shaukat2020survey}. Nevertheless, deep neural networks remain susceptible to adversarial perturbations---subtle, intentionally crafted modifications that induce misclassification.
\citep{szegedy2013intriguing} first demonstrated this vulnerability in image classifiers. Subsequent research first yielded the Fast Gradient Sign Method (FGSM) \citep{goodfellow2014explaining} and later introduced more advanced approaches such as Projected Gradient Descent (PGD) \citep{madry2017towards} and the Carlini \& Wagner (C\&W) attack \citep{carlini2017towards}, among others \citep{lapid2024open,lapid2022evolutionary,andriushchenko2020square,su2019one,lapid2023see,lapid2023patch}.

To safeguard these models, two defense paradigms have emerged. Adversarial robustness strengthens models via adversarial training \citep{madry2017towards,lapid2024fortify,andriushchenko2020understanding,wang2023better} or certified defenses \citep{li2023sok,cohen2019certified,chiangcertified,raghunathan2018certified}, while adversarial detection seeks to flag malicious inputs and reject them before they affect the model \citep{pang2022two,pinhasov2024xaibased,craighero2023unity,kazoom2025dont}. Early detectors applied input preprocessing or statistical tests \citep{feinman2017detecting,grosse2017statistical,xu2017feature,meng2017magnet}, and more recent methods exploit internal feature representations \citep{yeo2021robustness,xie2019feature,sun2024vitguard,mu2025robust}. 

In this work, we focus on adversarial detection, proposing an unsupervised method that flags malicious inputs without any attack labels. Unlike adversarial-training–based robustness—which requires re-annotation and retraining for each new threat—our approach is perturbation-agnostic and leaves the classifier untouched. \textbf{U-CAN} attaches lightweight contrastive heads to selected layers, thereby amplifying the separation between benign and adversarial inputs and achieving improved detection accuracy with minimal computational cost. 

Our key contributions are:
\begin{enumerate}
    \item \textbf{Unsupervised adversarial detection}. Our approach requires no labeled adversarial data, making it adaptable to various attack types without prior knowledge or specialized adversarial training.

    \item \textbf{No modifications to the target model}. External refinement heads handle detection, keeping the model’s architecture and weights unchanged—maintaining the original performance without any retraining.
    
    \item \textbf{Compatibility with existing methods}. The proposed framework--U-CAN integrates seamlessly with adversarial detection mechanisms that utilize intermediate model layers, further improving detection efficacy, as demonstrated in Section \ref{sec:results}.
\end{enumerate}

\section{Previous Work}
\label{sec:previous_work}

Adversarial attacks challenge the robustness of machine learning models, particularly in critical applications like image classification and object detection \citep{finlayson2019adversarial,guo2019simple,lapid2022evolutionary,alter2025on,carlini2017towards,lapid2023patch,lapid2023see,lapid2024open,carlini2017adversarial,vitracktamam2023foiling}. To counter these threats, adversarial detection methods have been widely explored. Many rely on adversarial examples during training, while others operate without them to improve generalization and reduce computational costs \citep{roth2019odds,li2017adversarial,dathathri2018detecting,ma2018characterizing,carrara2017detecting,metzen2022detecting}. This paper primarily focuses on unsupervised approaches, with a brief overview of supervised methods.

\subsection{Supervised Adversarial Detection}
\label{subsec:previous_supervised}

Supervised detectors learn from labeled adversarial examples to distinguish malicious from benign inputs. \citep{carrara2018adversarial} extract features from several network layers and, for each class, compute the distances between the input features and the corresponding benign-class medians; these distance sequences are then fed into an LSTM \citep{graves2012long} that classifies the input as adversarial or benign. \citep{lee2018simple} model class-wise feature activations as multivariate Gaussian distributions. They measure input deviations with Mahalanobis distances, and a logistic regression detector (trained on both benign and adversarial data) aggregates layer-wise distances to enhance detection reliability.

\subsection{Unsupervised Adversarial Detection}
\label{subsec:previous_unsupervised}

Unsupervised methods detect adversarial inputs without any training on labeled attack examples. This label-free strategy reduces complexity and improves generalization across attack types, as it requires neither retraining nor commitment to particular attack types, making unsupervised detection both challenging and a valuable research direction. Usually, such methods require setting an appropriate detection threshold. \citep{xu2017feature} introduce Feature Squeezing (FS), which applies transformations such as bit-depth reduction and spatial smoothing to simplify input features. Detection then is based on prediction discrepancies between the original and squeezed inputs. \citep{hendrycks2016baseline} propose Softmax Adversarial Detection (SAD)---a simple method which deems an input adversarial if its maximum softmax score falls below a chosen threshold. Because it uses only the classifier’s own confidences and no attack data, SAD is an unsupervised detector. \citep{papernot2018deep} introduce Deep k-NN (DKNN), which compares an input’s layer-wise features to those of benign calibration and training sets using $k$-nearest neighbors. The resulting votes yield p-values; Low p-values indicate low credibility of the input sample, which means anomaly or adversarial input. Deep Neural Rejection (DNR) \citep{sotgiu2020deep} enhances networks with auxiliary RBF-SVM classifiers at intermediate layers. An aggregator SVM combines their outputs to reject adversarial inputs with inconsistent layer-wise representations. This way, DNR effectively adds a “reject” neuron to the classifier. \cite{liu2022nowhere} propose AutoEncoder-based Adversarial Examples (AEAE), a shallow autoencoder trained only on benign data. It derives two metrics—reconstruction error and KL divergence between the classifier’s prediction on the original and reconstructed input—and feeds those metrics to an Isolation Forest \citep{liu2008isolation} that labels the input as benign or adversarial. \citep{chyou2023unsupervised} modify the classifier’s loss so that its raw logits also act as a detection score. Two KL-based terms push false-logits toward a uniform distribution while boosting the correct logit. To avoid a convergence to a one-hot output, adversarial examples ($\sim{10}\%$ of the training data) are injected, and only the false-logit loss is applied to them. At inference, an input is flagged as adversarial when its maximum or minimum logit crosses preset thresholds. Because \emph{some attack data is used during training}, the method is only \emph{partially} unsupervised. \citep{liu2022feature} present DCT Feature-Filter (DCTFF), which applies a discrete cosine transform to separate “dominant” (visible) coefficients from “recessive” ones often exploited by attacks. Coefficients below a threshold $\alpha$ are suppressed, and adversarial detection is performed by comparing the classifier’s predictions on the original and filtered images. \citep{app13064068} present ShuffleDetect (SD), which repeatedly permutes fixed-size image patches and re-evaluates the classifier. Since shuffling disrupts adversarial noise more than benign content, SD counts the fraction of shuffles whose predictions differ from the original; if this rate exceeds a preset threshold, the sample is flagged as adversarial.

\subsection{Key Differences from Prior Work}
Our methodology builds on intermediate representations for adversarial detection but introduces key differences:
\begin{enumerate}
    \item \textbf{No dependence on adversarial data.} Unlike supervised methods \citep{lee2018simple,carrara2018adversarial,chyou2023unsupervised}, which require adversarial examples for training, our approach is \emph{fully} unsupervised, eliminating computational overhead and bias from adversarial training.
    \item \textbf{Auxiliary networks with ArcFace-based feature refinement.} Rather than relying solely on raw feature statistics \citep{papernot2018deep,carrara2018adversarial,sotgiu2020deep}, we introduce lightweight auxiliary networks that refine feature spaces. These networks include: (i) a projection layer that maps features to a lower-dimensional space, and (ii) an ArcFace layer that enforces margin-based separation on a hypersphere, enhancing feature discrimination.
    \item \textbf{Layer-wise modularity for fine-grained detection.} Prior works often aggregate final-layer statistics \citep{lee2018simple,sotgiu2020deep}. In contrast, we integrate auxiliary networks across multiple layers ($L_{1}, L_{2}, \dots, L_{N}$), capturing diverse feature granularity and improving detection effectiveness without requiring supervised external classifiers such as LSTMs or logistic regression \citep{carrara2018adversarial,lee2018simple}.
    \item \textbf{Adversarial detection through a single forward pass.} At test time, unlike \cite{liu2022nowhere,xu2017feature,liu2022feature,app13064068}, our method requires only a single forward pass through the target model to both detect adversarial examples and perform its primary task (e.g., classification), rather than comparing multiple predictions to identify adversarial inputs.
\end{enumerate}

\section{Preliminaries}
\label{sec:preliminaries}
\textbf{Contrastive learning.} Contrastive learning \citep{oord2018representation,chen2020simple,khosla2020supervised} is a representation-learning paradigm \citep{botteghi2025unsupervised,fan2025scaling} that structures the latent space by \emph{pulling} semantically related samples together and \emph{pushing} unrelated samples apart. This process depends heavily on the training phase; accordingly, several contrastive strategies can be employed, e.g., a three-way strategy comprising \cite{schroff2015facenet}: (i) an \emph{anchor} example, (ii) a \emph{positive} example conveying the same semantic content (e.g., another augmentation of the same image), and (iii) a set of \emph{negative} examples drawn from the remainder of the mini-batch. A similarity function scores the affinity between any two feature vectors. The loss increases the similarity of each anchor–positive pair while reducing similarity to all anchor–negative pairs.

\textbf{ArcFace.} Originating in face-recognition research, ArcFace \citep{Deng_2022} loss adds a fixed angular margin to each class (or pseudo-label) logit before softmax, producing discriminative hyperspherical embeddings. The margin compacts same-class samples and widens inter-class gaps, while a scale factor parameter sharpens the output distribution. We apply these constraints at several intermediate layers, boosting feature separability which exposes adversarials.

\section{Methodology}
\label{sec:methodology}

\begin{figure*}[ht]
\centering
\includegraphics[trim={1cm 4cm 4.5cm 0.2cm},clip,scale=0.43]{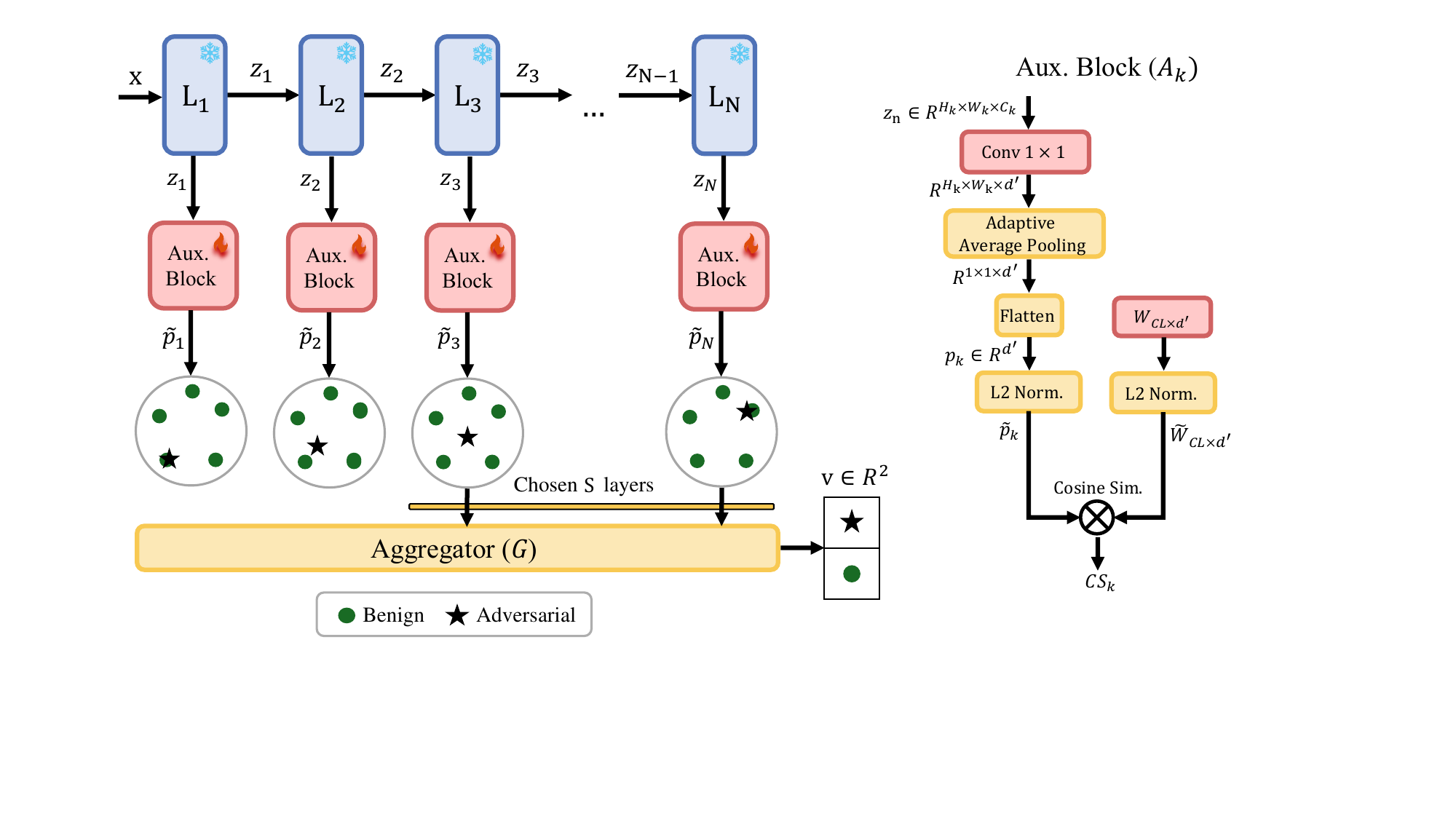}

\caption{Overview of our proposed method. The input $x$ passes through a frozen target model $\mathcal{M}$ with layers $\{L_1, L_2 \dots, L_N\}$, yielding features $\{\mathbf{z_1, z_2}, \dots, \mathbf{z_N}\}$. Each $\mathbf z_k$ is fed to an \textit{Aux. Block} $\mathcal{A}_k$ ($1{\times}1$ conv $\rightarrow$ adaptive-avg-pool $\rightarrow$ flatten $\rightarrow$ $\ell_2$-norm), producing refined vectors $\{\tilde{\mathbf{p}}_1, \tilde{\mathbf{p}}_1, \dots, \tilde{\mathbf{p}}_N\}$ that lie on unit hyperspheres anchored by the ArcFace learnable class centers $\mathbf{W_k} \in \mathbb{R}^{CL\times{d'}}$. Adversarial shifts (black stars) become more distinguishable from the well-separated benign clusters (green circles). An aggregator $\mathcal{G}$ is applied to the $S$ most informative auxiliaries---combines their outputs into an adversarial detection vector $\mathbf{v}$.}
\label{fig:algorithm}
\end{figure*}

We propose \emph{U-CAN}, an unsupervised adversarial detection framework built atop a frozen target network $\mathcal{M}$, which given an input $\mathbf{x} \in \mathbb{R}^{H \times W \times C}$, it extracts intermediate feature maps $\{\mathbf{z_k}\in \mathbb{R}^{H_k \times W_k \times C_k}\}_{k=1}^{N}$ for each layer $L_k\in \mathcal{M}$. As shown in Figure~\ref{fig:algorithm} and Algorithm~\ref{alg:ucan_shortest}, each $\mathbf{z_k}$ is refined by an \emph{Aux.~Block} ($\mathcal{A}_k$)—comprising a $1\!\times\!1$ convolution, adaptive average pooling, flattening, and $\ell_{2}$-normalization—followed by an \emph{ArcFace} layer, yielding a well-separated embedding space. Finally, an \emph{Aggregator} ($\mathcal{G}$) combines these refined embeddings into a compact detection vector $\mathbf{v}\in\mathbb{R}^2$, enabling robust adversarial detection. The auxiliary networks transform intermediate feature representations into a more structured space, where adversarial perturbations become more distinguishable. These refined representations can be integrated into various feature aggregation and anomaly detection methods.  For simplicity, this paper focuses on classification, yet the approach extends to other tasks and domains—for instance, treating object-recognition classes as per-instance pseudo-labels or refining token features in language models.

\subsection{Auxiliary Networks}
Each chosen layer $L_k$ is associated with an Aux. Block, $\mathcal{A}_k$, which refines feature representations to amplify adversarial discrepancies. Each auxiliary block consists of projection and ArcFace \citep{Deng_2022} components.

\paragraph{(1) Projection component.} This module projects the feature maps $\{\mathbf{z}_k\}$ and flattens them into fixed, compact representations $\{\mathbf{p}_k\in \mathbb{R}^{d'}\}_{k=1}^{N}$ where $d' \ll H_k \cdot W_k \cdot C_k$. The projection consists of a $1\times1$ convolution to reduce channels, followed by adaptive average pooling.

\paragraph{(2) ArcFace linear layer.} While training, each transformed feature vector $\mathbf{p}_k$ is processed through  an ArcFace-based transformation \citep{Deng_2022}:
\begin{equation}
    \tilde{\mathbf{W}}_k = \normalize{\mathbf{W}_k}, \quad \tilde{\mathbf{p}}_k = \normalize{\mathbf{p}_k}; \quad CS_k = \tilde{\mathbf{W}}_k \tilde{\mathbf{p}}_k,
\end{equation}
where $\mathbf{W}_k \in \mathbb{R}^{CL \times d'}$ are the learnable ArcFace weight matrices representing class centers, $CL$ is the number of classes, and $CS_k \in \mathbb{R}^{CL}$ is the cosine similarity scores between the normalized projected feature vector $\tilde{\mathbf{p}}_k$ and the normalized class centers $\tilde{\mathbf{W}}_k$. ArcFace \citep{Deng_2022} projects features onto a unit hypersphere by adding a marginal penalty while training, widening inter-class gaps while tightening intra-class clusters. Consequently, vectors on the auxiliary ArcFace hyperspheres $\tilde{\mathbf{p}}_k$ become highly sensitive to small perturbations: minor shifts become more pronounced and less noisy in the refined feature spaces, making it easier to expose adversarial examples through the target model layers.

\subsection{Training Procedure}
Our framework does not require adversarial samples for training. Given a pretrained target model $\mathcal{M}$, each auxiliary network $\mathcal{A}_k$ is trained to refine embeddings while keeping $\mathcal{M}$ frozen. During training, the objective is to maximize intra-class similarity while enforcing a margin that separates instances from different classes. The $k$-th auxiliary ArcFace \citep{Deng_2022} loss can be formulated as:
\begin{equation}
    \mathcal{L}_k = -\log \frac{s \cdot e^{\cos(\theta^{k}_{y_i} + m)}}{s \cdot e^{\cos(\theta^{k}_{y_i} + m)} + \sum_{j=1, j \neq i}^{CL} s \cdot e^{\cos(\theta^{k}_{y_j})}},
\end{equation}
where $m$ is the angular margin, $s$ is the hypersphere scale factor, $y_i$ is the target class, $CL$ is the number of classes and $\theta^{k}_{y_i}$ is the angle between the feature vector and the corresponding class center at feature level $k$. Thus, the global loss function is computed as:
\begin{equation}
    \mathcal{L}_{global} = \frac{1}{N} \sum_{i=1}^{N}\mathcal{L}_k.
\end{equation}

where $N$ is the number of layers.

\subsection{Layer-Wise Fusion}
Refined feature embeddings $\tilde{\mathbf{p}}_k$ from different layers are fused to improve adversarial detection:
\begin{equation}
    \mathbf{v} = \mathcal{G}(\{\tilde{\mathbf{p}}_s\}) \mid s\in\mathcal{S},
\end{equation}
where $\mathcal{G}$ is a given feature aggregation algorithm and $\mathcal{S}$ ($|S| \leq N)$ denotes the subset of auxiliary blocks whose informative feature representations contribute most to the final adversarial-detection score.

\paragraph{Inference.} During inference, a test sample $\mathbf{x}$ is processed through $\mathcal{M}$, producing feature maps $\mathbf{z}_k$. Each selected $\mathbf{z}_s$ is refined by its corresponding auxiliary block, generating $\tilde{\mathbf{p}}_s$. These embeddings are fused and analyzed through $\mathcal{G}$ to determine whether the input is adversarial based on its deviation from the learnable benign feature distributions. The aggregator can use either the normalized representations ($\tilde{\mathbf{p}}_s$) or their logits ($CS_s$) relative to the learnable class centers ($\mathbf{W}_s$), depending on its specific algorithm.

\begin{algorithm}
\scriptsize
\caption{U-CAN}
\label{alg:ucan_shortest}
\SetKwInOut{Input}{Input}
\SetKwInOut{Output}{Output}

\Input{Frozen model $\mathcal{M}$ with layers $L_1,\ldots,L_N$, benign data and off-the-shelf aggregator $\mathcal{G}$.}
\Output{Trained adversarial detector $\mathcal{D}$.}

\BlankLine
\textbf{1.} Initialize Aux.\ Blocks $\{\mathcal{A}_k\}_{k=1}^N$.

\BlankLine
\textbf{2.} Stack $\{\mathcal{A}_k\}_{k=1}^N$ on top of $\mathcal{M}$ and train them simultaneously using $\mathcal{L}_{global}$.

\BlankLine
\textbf{3.} Compute validation score $CS^{(avg)}_k$ per $\mathcal{A}_k$; select the best $\mathcal{S}$ blocks $\{\,\mathcal{A}_s \mid s \in \mathcal{S}\}$.

\vspace{5pt}

\textbf{Inference:} \\ 
\For{test sample $\mathbf{x}$}{
  Feed $\textbf{x}$ through $\mathcal{M}$. \\  
  Extract embeddings from $\{\,\mathcal{A}_s\}_{s\in\mathcal{S}}$; feed into $\mathcal{G}$.\\
  Flag $\mathbf{x}$ as adversarial or benign.
}

\end{algorithm}

\vspace{-0.2cm}
\section{Experimental Setting}\label{sec:experimental_setting}

\subsection{Datasets}
\label{datasets_section}
We conducted our experiments on three datasets: CIFAR-10 \citep{krizhevsky2009learning}, Mammals \citep{asaniczka_2023}, and a subset of ImageNet \citep{russakovsky2015imagenet}. These choices ensure a variety of classes, resolutions, and receptive fields while remaining computationally feasible. CIFAR-10 contains 60,000 images of size 32x32 across 10 classes, Mammals includes 13,751 images of size 256x256 spanning 45 classes, and our ImageNet subset focuses on 6 randomly selected classes (1,050 images per class).

For each dataset, we split the data into training, validation, calibration, and test sets. The training and validation sets were used to train both the baseline classifiers and our auxiliary networks. The calibration set stored reference feature vectors for the DKNN \citep{papernot2018deep} adversarial detection method, and the final test set contained benign examples later transformed into adversarial queries.

\subsection{Models}
\label{models_section}
To accommodate diverse architectural styles, we used three classification models: ResNet-50 \citep{he2016deep}, VGG-16 \citep{simonyan2014very}, and ViT-B-16 \citep{dosovitskiy2020image}. ResNet-50 and VGG-16 are prominent CNNs with different depths and layer structures, whereas ViT-B-16 is based on the transformer paradigm, representing a more recent approach to image classification.

We first trained 9 classifiers (one for each dataset-model combination) for 250 epochs, initializing from ImageNet-pretrained weights \citep{russakovsky2015imagenet} and selecting the best checkpoints by F1 validation scores. We then constructed and trained our auxiliary networks atop these frozen classifiers for up to 1,500 epochs. During auxiliary-network training, we measured inter-class separation and intra-class cohesion using the \emph{total cosine similarity} (TCS), calculated on the validation set as the average across classes and layers of the positive-class similarity minus the negative-class similarities (Equation \ref{eq5a}, \ref{eq5b}).

\begin{raggedleft}
\begin{minipage}{1.0\linewidth}
\footnotesize
\begin{subequations}
\label{eq5}
\begin{align}
    & CS^{+}_k = \tilde{\mathbf{W}}^{y_i^{T}}_k \tilde{\mathbf{p}}_k;\quad
    CS^{-}_k = \frac{1}{CL-1} \sum_{j=1,j\neq{i}}^{CL}{\tilde{\mathbf{W}}^{y_j^{T}}_k 
    \tilde{\mathbf{p}}_k}
    \label{eq5a} \\
    & CS^{(avg)}_k = \frac{1}{2} (CS^{+}_k - CS^{-}_k);\quad
    TCS = \frac{1}{N} \sum_{i=1}^{N}{CS^{(avg)}_k}
    \label{eq5b}
\end{align}
\end{subequations}
\end{minipage}
\end{raggedleft}

where $y_i$ is the positive class, $\tilde{\mathbf{W}}^{y_i^{T}}_k$ denotes the transposed normalized vector for the positive class center, $\tilde{\mathbf{W}}^{y_{j\neq{i}}^{T}}_k$ are the transposed normalized vectors for the other (negative) classes, and $\tilde{\mathbf{p}}_k$ is the normalized feature vector of the current sample. We used \texttt{AdamW} with \texttt{ReduceLROnPlateau} and a learning rate of $1\times 10^{-5}$ for CIFAR10, and $1\times 10^{-4}$ for both the ImageNet subset and Mammals. Data augmentations included random zoom, brightness, hue, saturation, rotation, a small random perspective transform, and random box masking.

\begin{figure}[ht!]
\centering

\begin{subfigure}[b]{0.8\linewidth}
    \centering
    \includegraphics[trim={0.5cm 0.5cm 0.5cm 0.5cm}, 
                     width=1\linewidth]{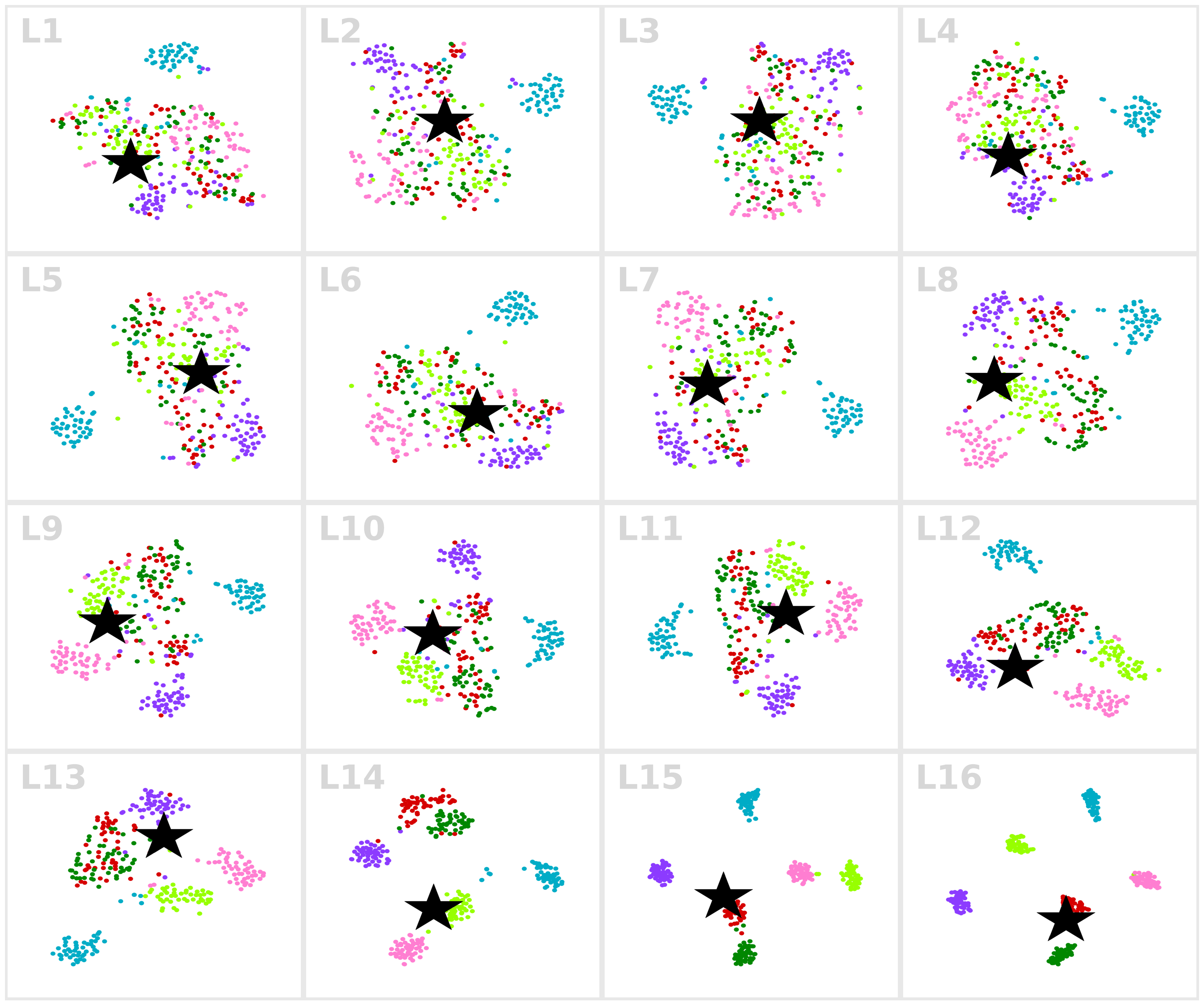}
    \label{fig:vis_feats_org-resnet50}
\end{subfigure}

\vspace{-0.3em} 

\begin{subfigure}[b]{0.8\linewidth}
    \centering
    \includegraphics[trim={0.5cm 0.5cm 0.5cm 0.5cm}, 
                     width=1\linewidth]{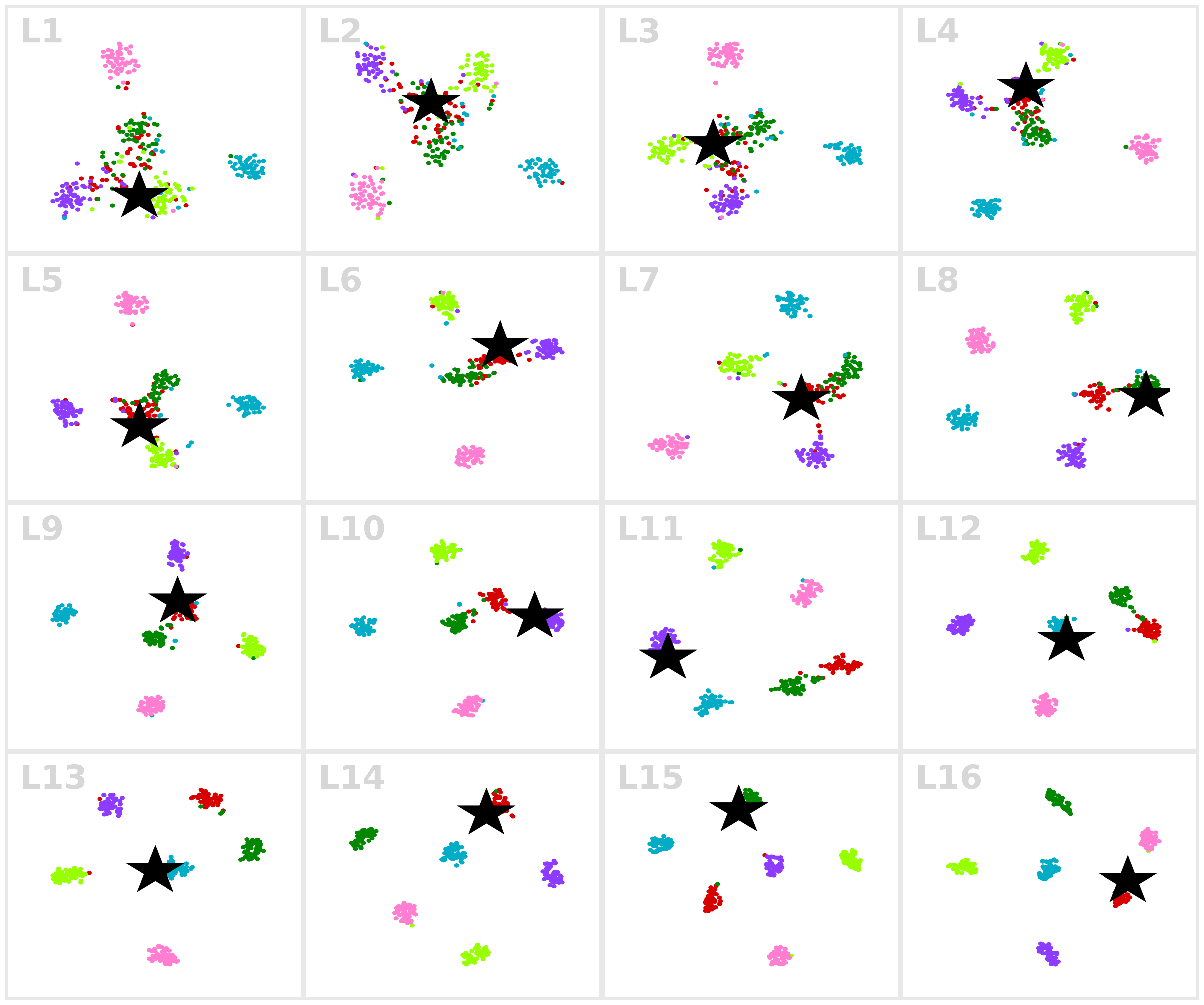}
    \includegraphics[trim={0.5cm 1.8cm 0.5cm 0.5cm},scale=0.2]{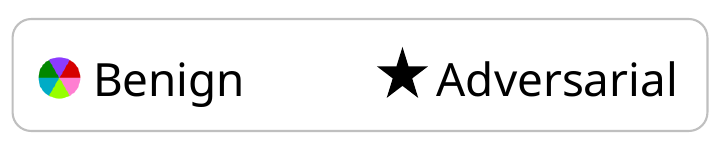}
    \label{fig:vis_feats_norm-resnet50}
\end{subfigure}

\vspace{0.2cm}

\caption{Layer-wise t-SNE-reduced feature visualizations of ResNet-50 on ImageNet validation set: \textbf{Top}--raw features $\{\mathbf{z}_n\}_{1}^{16}$; \textbf{Bottom}--U-CAN’s contrastive features $\{\tilde{\mathbf{p}}_n\}_{1}^{16}$. From top-left ($L_0$) to bottom-right ($L_{16}$), each plot shows the benign class clusters (colors) and a single adversarial sample (black star). Without U-CAN, adversarial points blend in; with U-CAN, refined features sharpen class boundaries, exposing adversarial crossings.}

\label{fig:vis_feats_merged}
\end{figure}

\subsection{Attacks}
\label{attacks_definition}
Below, we summarize key attack methods evaluated in our work in the following experiments against adversarial detection methods \citep{papernot2018deep,sotgiu2020deep,xu2017feature,hendrycks2016baseline,liu2022nowhere,liu2022feature,app13064068}.
\begin{enumerate}
    \item \textbf{PGD (Projected Gradient Descent)} \citep{madry2017towards} generates adversarial perturbations, iteratively using the gradient sign of the loss function, while projecting each step back into an imperceptible perturbation space.
    \item \textbf{C\&W (Carlini \& Wagner)} \citep{carlini2017towards} formulates adversarial example generation as an optimization problem, minimizing perturbations while ensuring misclassification.
    \item \textbf{AA (AutoAttack)} \citep{croce2020reliable} is an ensemble of multiple attacks, including Auto-PGD, FAB \citep{croce2020minimally}, and Square Attack \citep{andriushchenko2020square}, designed to identify model vulnerabilities through diverse attack strategies.
\end{enumerate}

For attacks parameterized by an $\epsilon$, we evaluate at two $\ell_\infty$ budgets—$8/255$ and $16/255$. All iterative attacks run for 200 steps, except those inside AutoAttack (AA) \citep{croce2020reliable}, which use 100 steps per sub-attack. For C\&W \citep{carlini2017towards} we use $c=0.5$, $\kappa=0$, and a learning rate of $10^{-3}$.

\subsection{Adaptive Attacks}
\label{adaptive_attacks_definition}
We evaluated \emph{adaptive attacks} against the four strongest and SotA baseline detectors and against our best method variant U-CAN+DNR according to Table \ref{tab:exp-results}.
\begin{enumerate}
  \item \textbf{ADA-DKNN} \citep{sitawarin2019robustness}: 
        Approximates the gradients of the $k$-NN vote at intermediate feature layers and drives adversarial examples toward benign representations belonging to the closest incorrect class.  
        Because DNR \citep{sotgiu2020deep} also relies on the class separability of these feature spaces, ADA-DKNN is expected to degrade the performance of both DKNN \citep{papernot2018deep} and DNR-based \citep{sotgiu2020deep} detectors.
  \item \textbf{Adaptive AEAE (ADA-AEAE)}:  
        A specific attack we devised for AEAE \citep{liu2022nowhere}, based on the C\&W attack \citep{carlini2017towards}. It simultaneously reduces: 
        (i) the auto-encoder reconstruction error of the perturbed input, and (ii) the KL divergence between the classifier’s predictions on the adversarial input and on its reconstruction, thereby neutralizing AEAE’s detection cue.
  \item \textbf{Adaptive DCTFF (ADA-DCTFF)}:
        A specific attack we devised for \citep{liu2022feature} based on the $\ell_{2}$–C\&W attack \citep{carlini2017adversarial}. ADA-DCTFF adds a DCTFF‑adaptivity term, weighted by~$\mu$, that uses the square-root of the Jensen–Shannon Divergence (JSD) \citep{lin2002divergence}--a symmetric version of the KL divergence, to enforce consistency between the classifier’s logits on the original and DCT‑filtered images under perturbation. At each optimization step, we uniformly sample $m$ cutoff fractions $\alpha\in[\alpha_{\min},\alpha_{\max}]$ and apply a soft sigmoid mask—whose slope is linearly annealed from $\beta_{\min}$ to $\beta_{\max}$—to suppress low‑frequency DCT coefficients, while ensuring differentiability along with the JSD.
\newline
  \item \textbf{Adaptive Shuffle Detect (ADA-SD)}:
        A simple variant of the C\&W attack \citep{carlini2017adversarial} that jointly minimizes the confidence scores of both the target classifier and the Shuffle Detect adversarial detector \citep{app13064068}. ADA-SD assumes knowledge of the patch permutations, which is valid under a white-box setting. To obtain a differentiable binary (adversarial/benign) confidence measure between 0 and 1, we use the square-root Jensen–Shannon Divergence \citep{lin2002divergence} between predictions on the original input and its permutations.

\end{enumerate}

For ADA-DKNN, we kept the hyperparameters of \citep{sitawarin2019robustness}, except that we set the iteration count to 400 and tuned $m$ to $100$--the number of nearest target-class samples used in the k-NN approximation. For ADA-AEAE, we used $c = 5 \times 10^{-5}$ and the same two $\epsilon$ values as above ($16/255$ and $8/255$). For ADA-DCTFF we used $(\alpha_{\min},\alpha_{\max})=(0.05, 0.95)$ with $m=10$ and $(\beta_{\min},\beta_{\max})=(10, 40)$, with adaptive loss weight $\mu=5$. Lastly, in ADA-SD, we used the same C\&W hyperparameters, with an adaptive loss weight of $30$.

\subsection{Selecting the Subset of Auxiliary Networks}
To determine the appropriate layer indices of auxiliary networks before conducting each experiment, we calculated $CS^{(avg)}_k$ from Equation \ref{eq5b} on the validation set for each feature level of the auxiliary networks. We then chose the best \(\mathcal{S}\) layers with the highest $\{CS^{(avg)}_k\}^N_{k=1}$ scores for each model and dataset. Figure \ref{fig:vis_feats_merged} provides a reduced view comparison of our refined U-CAN features and the original ones for the ResNet-50 backbone. Specifically, for each experiment, we selected layers from an offset $s \in \{1,\dots,N\}$ onward—$\{L_k\}_{k=s}^N$—ensuring sufficient inter-class separation and intra-class clustering. The exact $s$ values used in each experiment are provided in \ref{tab:layer-offsets}.

\subsection{Overhead Analysis}
\label{sec:overhead}
 In this section, we provide complete details on the hyperparameters budget, memory overhead, and computational cost.

\paragraph{Hyperparameters budget.}
U‑CAN introduces only three hyperparameters per auxiliary block: ArcFace scale $s$, margin $m$--both fixed to the defaults in \citep{Deng_2022}, and a balanced latent dimension $d'$ set once per backbone (512/256/768 for ResNet‑50, VGG‑16, ViT‑B-16). We chose reasonable blocks to attach the auxiliary networks 16, 13, and 12, respectively. 

\paragraph{Parameter and memory overhead.}
Table \ref{tab:params} reports the additional parameters and VRAM introduced
by the auxiliary networks, together with their percentage of the full model. The overhead is below 1.2\% for ViT and VGG backbones and $\approx$25\% for ResNet-50, due to its smaller main linear head—compared to the other models—and the larger number of selected auxiliary blocks.

\begin{table}[t]
\centering

\makebox[\linewidth]{%
\begin{minipage}[t]{.55\linewidth}
  \captionof{table}{Parameter overhead of\\the auxiliary networks.}
  \label{tab:params}
  \begingroup
    \setlength{\tabcolsep}{3pt}
    \renewcommand{\arraystretch}{0.9}
    \footnotesize
    \resizebox{\linewidth}{!}{%
      \begin{tabular}{@{}llr@{}}
        \toprule
        \textbf{Backbone} &
        \shortstack[c]{\textbf{Aux.}\\\textbf{params}} &
        \shortstack[c]{\textbf{\%}\\\textbf{of total}}\\
        \midrule
        ViT‑B/16 (Mammals)     & 1.01M & 1.16 \\
        ViT‑B/16 (ImageNet‑6)  & 0.65M & 0.75 \\
        ViT‑B/16 (CIFAR‑10)    & 0.68M & 0.79 \\
        ResNet‑50 (Mammals)    & 8.10M & 25.6 \\
        ResNet‑50 (ImageNet‑6) & 7.78M & 24.9 \\
        ResNet‑50 (CIFAR‑10)   & 7.82M & 24.9 \\
        VGG‑16 (Mammals)       & 1.23M & 1.00 \\
        VGG‑16 (ImageNet‑6)    & 1.10M & 0.90 \\
        VGG‑16 (CIFAR‑10)      & 1.11M & 0.91 \\
        \bottomrule
      \end{tabular}}
  \endgroup
\end{minipage}%
\hfill 
\begin{minipage}[t]{.4\linewidth}
  \captionof{table}{Chosen $s$ layer\\offsets for experiments.}
  \label{tab:layer-offsets}
  \begingroup
    \setlength{\tabcolsep}{0pt}
    \renewcommand{\arraystretch}{0.9}
    \footnotesize
    \vspace{-0.5pt}
    \resizebox{0.96\linewidth}{!}{%
      \begin{tabular}{@{}lc@{}}
        \toprule
        \textbf{Backbone} & \shortstack[c]{\textbf{Layer}\\\textbf{offset}}\\
        \midrule
        ViT‑B/16 (Mammals)     & 6 \\
        ViT‑B/16 (ImageNet‑6)  & 3 \\
        ViT‑B/16 (CIFAR‑10)    & 5 \\
        ResNet‑50 (Mammals)    & 7 \\
        ResNet‑50 (ImageNet‑6) & 3 \\
        ResNet‑50 (CIFAR‑10)   & 5 \\
        VGG‑16 (Mammals)       & 8 \\
        VGG‑16 (ImageNet‑6)    & 5 \\
        VGG‑16 (CIFAR‑10)      & 6 \\
        \bottomrule
      \end{tabular}}
  \endgroup
\end{minipage}%
}
\end{table}
  
\paragraph{Computational cost.}
We evaluate computational cost by running all adversarial detection methods
\citep{xu2017feature,hendrycks2016baseline,papernot2018deep,sotgiu2020deep,
liu2022nowhere,liu2022feature,app13064068}, for each model and dataset, on an
Intel(R) Xeon(R) E5-1650 v4 @ 3.60 GHz CPU with a single thread and 128 GB
RAM, since several implementations cannot execute on a GPU. Table \ref{tab:latency} reports the inference computational cost of all methods.

\paragraph{Take-away.} U-CAN adds negligible storage and latency on transformer and VGG backbones and
remains modest on ResNet-50, while delivering the detection gains reported in
\autoref{sec:results}. Because inference still involves a single forward pass—and the refined features are, in some models, more compact than the full set of original layer-wise features—U-CAN meets the low-latency requirements of safety-critical applications.

\section{Results}
\label{sec:results}

{

\sisetup{
  round-mode      = places ,
  round-precision = 3
}

\begin{table*}[ht!]
\centering
\caption{Inference runtime cost (seconds) of each adversarial detector for all experiments (batch size $8$, averaged over $10$ iterations). The \textbf{boldface} entry denotes the lowest average (in seconds), and the \underline{underlined} entry denotes the second lowest.}
\label{tab:latency}

{\scriptsize
\setlength{\tabcolsep}{2pt}
\renewcommand{\arraystretch}{}
\resizebox{0.78\textwidth}{!}{%
\begin{tabular}{
  l|l|*{7}{C{1.0cm}}|G{1.8cm}|G{1.8cm}}
\toprule
\multicolumn{1}{c|}{\textbf{Dataset}} &
\multicolumn{1}{c|}{\textbf{Model}} &
FS & SAD & DKNN & DNR &
AEAE & DCTFF & SD &
\textbf{U-CAN+DKNN} & \textbf{U-CAN+DNR}\\
\midrule
\multirow[c]{3}{*}{\textbf{CIFAR-10}} &
ResNet-50 & \num{0.297} & \num{0.046} & \num{1.284} & \num{0.103} & \num{0.098} & \num{0.109} & \num{4.667} & \num{1.013} & \num{0.075} \\
& VGG-16   & \num{0.297} & \num{1.071} & \num{1.188} & \num{0.093} & \num{2.462} & \num{0.152} & \num{6.763} & \num{0.348} & \num{0.047} \\
& ViT      & \num{4.994} & \num{1.134} & \num{2.256} & \num{3.398} & \num{7.215} & \num{2.290} & \num{113.995} & \num{1.536} & \num{1.138} \\
\midrule
\multirow[c]{3}{*}{\textbf{Mammals}} &
ResNet-50 & \num{1.867} & \num{0.350} & \num{0.808} & \num{1.003} & \num{2.305} & \num{0.739} & \num{35.377} & \num{0.713} & \num{0.566} \\
& VGG-16   & \num{4.753} & \num{1.071} & \num{1.760} & \num{2.898} & \num{6.420} & \num{2.176} & \num{108.206} & \num{1.480} & \num{1.436} \\
& ViT      & \num{4.952} & \num{1.202} & \num{1.347} & \num{3.364} & \num{7.133} & \num{2.276} & \num{113.202} & \num{1.194} & \num{1.148} \\
\midrule
\multirow[c]{3}{*}{\textbf{ImageNet}} &
ResNet-50 & \num{1.726} & \num{0.350} & \num{0.715} & \num{0.994} & \num{2.340} & \num{0.757} & \num{35.461} & \num{0.722} & \num{0.568} \\
& VGG-16   & \num{4.753} & \num{1.042} & \num{1.126} & \num{2.905} & \num{6.495} & \num{2.123} & \num{105.292} & \num{1.464} & \num{1.416} \\
& ViT      & \num{4.861} & \num{1.134} & \num{1.285} & \num{3.333} & \num{7.101} & \num{2.303} & \num{114.602} & \num{1.194} & \num{1.126} \\
\midrule
\multicolumn{2}{c|}{\textbf{Average}} &
\num{3.129} & \textbf{\num{0.701}} & \num{1.459} & \underline{\num{0.804}} &
\num{1.643} & \num{1.436} & \num{70.841} &
\num{1.075} & \num{0.836}\\
\bottomrule
\end{tabular}}}
\end{table*}
}

\begin{figure}[htb]
\raggedright            

\begin{minipage}{0.8\linewidth}  
  \includegraphics[trim=0.5cm 0.5cm 0.5cm 0.5cm,width=\linewidth]
                   {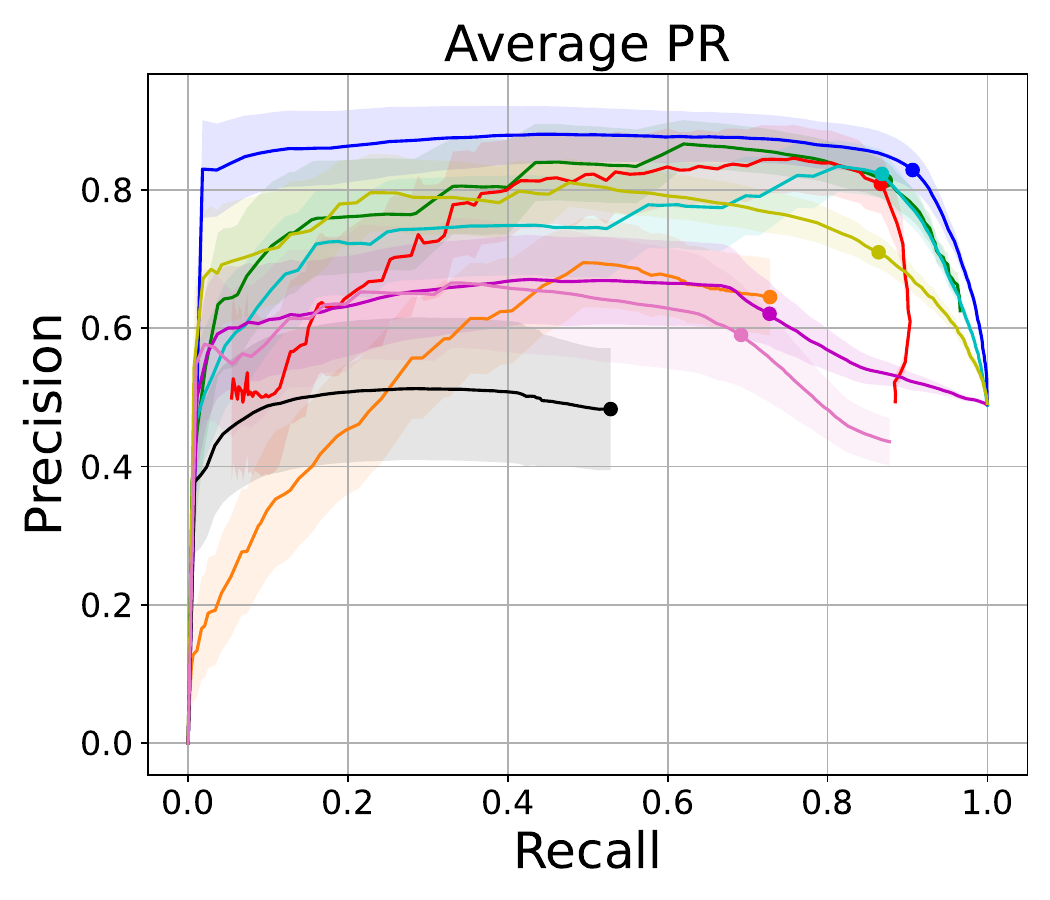}

  \vspace{1.2mm}
  \hspace{3mm}
  \includegraphics[trim=0.5cm 0.5cm 0.5cm 0.5cm,width=\linewidth]
                   {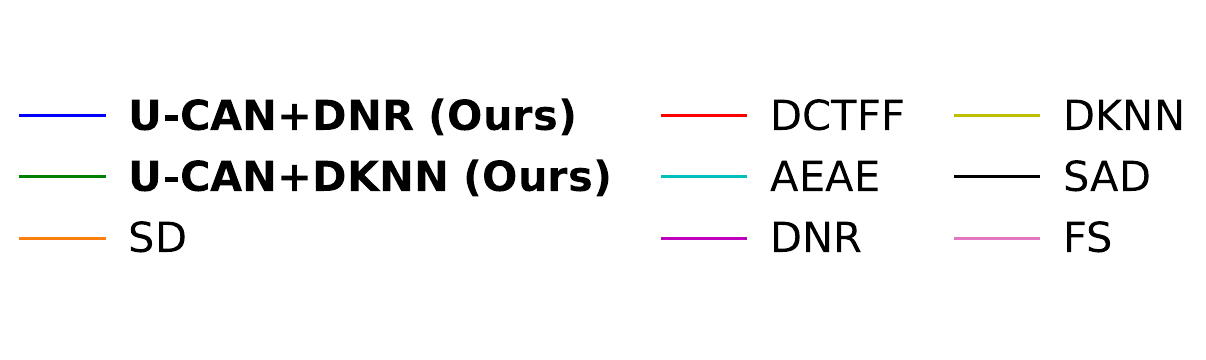}

  \vspace{-3mm}
  \caption{Average precision‑recall curves for each method on all datasets,
           models, attacks, and $\epsilon$ values. The thicker point
           marks the best F1, the transparent band is the scaled variance.}
  \label{fig:average_PR}
\end{minipage}

\end{figure}

\paragraph{Comparative evaluation.}
We benchmarked U-CAN, combined with DKNN and DNR (U-CAN+DKNN / U-CAN+DNR), against seven fully unsupervised detectors—FS, SAD, DKNN, DNR, AEAE, DCTFF and SD \citep{xu2017feature,hendrycks2016baseline,papernot2018deep,sotgiu2020deep,liu2022nowhere,liu2022feature,app13064068}. However, all methods require detection threshold tuning. For each method–attack–$\epsilon$–model–dataset combination, we generated Precision–recall (PR) curves and chose the threshold that maximized F1; the resulting scores are listed in Table \ref{tab:exp-results}. In \autoref{fig:average_PR}, we present a compact summary of the numerous PR curves: Averaged PR curves were generated for each detection method across all experiments (models, datasets, attacks, and $\epsilon$ values).

Overall, our method outperforms all other compared adversarial‑detection methods on average, achieving the \textbf{highest} F1 score of $90.13\%$ with the U‑CAN+DNR variant, improving the original DNR \citep{sotgiu2020deep} by \textcolor{Green}{$11.8\%$} while maintaining approximately the same inference latency. These findings demonstrate that refining the target model’s features makes U‑CAN highly effective for adversarial detection.

\begin{table*}[ht!]
\centering
\caption{F1-scores of all compared adversarial detectors:  
FS~\citep{xu2017feature}, SAD~\citep{hendrycks2016baseline}, DKNN~\citep{papernot2018deep},  
DNR~\citep{sotgiu2020deep}, AEAE~\citep{liu2022nowhere}, DCTFF \citep{liu2022feature}, SD \citep{app13064068}, and our
\textbf{U-CAN+DKNN} and \textbf{U-CAN+DNR} variants on all experiment combinations attack-$\epsilon$-dataset-model. The \textbf{boldface} entry denotes the best score, the \underline{underlined} entry the second best, and the \textcolor{Green}{green} text indicates the gain from stacking other detectors on U-CAN.}
\vspace{-4pt}
\label{tab:exp-results}

{\scriptsize
\setlength{\tabcolsep}{2pt}
\renewcommand{\arraystretch}{}
\resizebox{0.78\textwidth}{!}{%
\begin{tabular}{
  l|l|A{1.7cm}|*{7}{C{1cm}}|G{1.8cm}|G{1.8cm}}
\toprule
\multicolumn{1}{c|}{\textbf{Dataset}} &
\multicolumn{1}{c|}{\textbf{Model}}  &
\multicolumn{1}{c|}{\textbf{Attack}} &
FS & SAD & DKNN & DNR &
AEAE & DCTFF & SD &
\textbf{U-CAN+DKNN} & \textbf{U-CAN+DNR}\\
\midrule
\multirow[c]{21}{*}{\vspace{-12pt}\textbf{CIFAR-10}}
 & \multirow[c]{7}{*}{\vspace{-3pt}ResNet-50}
 & $\text{PGD}_{\epsilon=16/255}$ & \num{89.70} & \num{81.90} & \num{75.80} & \num{66.50} & \num{99.10} & \num{78.65} & \num{67.15} & \num{85.50} & \num{88.50}\\
 &  & $\text{PGD}_{\epsilon=8/255}$  & \num{90.50} & \num{76.50} & \num{77.80} & \num{66.50} & \num{90.30} & \num{81.60} & \num{67.85} & \num{84.80} & \num{85.50}\\
 &  & $\text{C\&W}_{\epsilon=16/255}$ & \num{83.70} & \num{66.50} & \num{70.50} & \num{67.80} & \num{84.80} & \num{68.62} & \num{65.52} & \num{81.20} & \num{85.40}\\
 &  & $\text{C\&W}_{\epsilon=8/255}$  & \num{84.20} & \num{66.50} & \num{72.10} & \num{67.00} & \num{76.50} & \num{73.21} & \num{66.34} & \num{84.90} & \num{84.30}\\
 &  & $\text{AA}_{\epsilon=16/255}$   & \num{84.30} & \num{66.50} & \num{69.90} & \num{67.20} & \num{94.30} & \num{74.46} & \num{63.16} & \num{75.30} & \num{77.40}\\
 &  & $\text{AA}_{\epsilon=8/255}$    & \num{85.60} & \num{66.50} & \num{71.10} & \num{67.10} & \num{78.00} & \num{74.38} & \num{65.66} & \num{78.30} & \num{77.00}\\ \cmidrule{2-12}
 & \multirow[c]{7}{*}{\vspace{-3pt}VGG-16}
 & $\text{PGD}_{\epsilon=16/255}$ & \num{74.90} & \num{60.70} & \num{81.50} & \num{75.40} & \num{98.00} & \num{76.12} & \num{69.65} & \num{89.30} & \num{90.50}\\
 &  & $\text{PGD}_{\epsilon=8/255}$  & \num{78.70} & \num{64.80} & \num{81.00} & \num{68.80} & \num{81.70} & \num{76.12} & \num{71.22} & \num{87.80} & \num{91.20}\\
 &  & $\text{C\&W}_{\epsilon=16/255}$ & \num{66.40} & \num{0.00} & \num{73.90} & \num{75.40} & \num{83.70} & \num{61.65} & \num{66.32} & \num{70.90} & \num{74.00}\\
 &  & $\text{C\&W}_{\epsilon=8/255}$  & \num{66.50} & \num{0.00} & \num{73.70} & \num{69.50} & \num{70.60} & \num{62.34} & \num{66.67} & \num{75.90} & \num{73.80}\\
 &  & $\text{AA}_{\epsilon=16/255}$   & \num{66.50} & \num{0.00} & \num{69.50} & \num{77.40} & \num{89.50} & \num{65.86} & \num{65.81} & \num{64.50} & \num{80.00}\\
 &  & $\text{AA}_{\epsilon=8/255}$    & \num{66.50} & \num{0.00} & \num{72.40} & \num{72.20} & \num{70.20} & \num{63.86} & \num{67.01} & \num{69.40} & \num{73.40}\\ \cmidrule{2-12}
 & \multirow[c]{7}{*}{\vspace{-3pt}ViT}
 & $\text{PGD}_{\epsilon=16/255}$ & \num{96.50} & \num{97.20} & \num{89.90} & \num{69.00} & \num{99.50} & \num{95.65} & \num{84.65} & \num{94.90} & \num{97.90}\\
 &  & $\text{PGD}_{\epsilon=8/255}$  & \num{95.20} & \num{96.70} & \num{88.50} & \num{68.30} & \num{99.20} & \num{96.19} & \num{83.01} & \num{96.20} & \num{97.10}\\
 &  & $\text{C\&W}_{\epsilon=16/255}$ & \num{94.60} & \num{96.80} & \num{86.40} & \num{66.60} & \num{98.70} & \num{93.99} & \num{81.49} & \num{93.10} & \num{95.90}\\
 &  & $\text{C\&W}_{\epsilon=8/255}$  & \num{94.30} & \num{97.00} & \num{86.60} & \num{66.70} & \num{98.70} & \num{95.38} & \num{81.49} & \num{94.00} & \num{95.30}\\
 &  & $\text{AA}_{\epsilon=16/255}$   & \num{93.90} & \num{96.20} & \num{86.60} & \num{66.50} & \num{98.70} & \num{95.56} & \num{74.16} & \num{93.00} & \num{96.50}\\
 &  & $\text{AA}_{\epsilon=8/255}$    & \num{93.90} & \num{96.20} & \num{84.10} & \num{66.70} & \num{98.70} & \num{96.79} & \num{77.83} & \num{91.40} & \num{95.00}\\
\midrule
\multirow[c]{21}{*}{\vspace{-12pt}\textbf{Mammals}}
 & \multirow[c]{7}{*}{\vspace{-3pt}ResNet-50}
 & $\text{PGD}_{\epsilon=16/255}$ & \num{91.30} & \num{67.90} & \num{89.00} & \num{99.00} & \num{99.70} & \num{97.61} & \num{95.74} & \num{94.30} & \num{97.60}\\
 &  & $\text{PGD}_{\epsilon=8/255}$  & \num{94.00} & \num{77.00} & \num{83.80} & \num{98.80} & \num{99.60} & \num{97.74} & \num{95.27} & \num{91.60} & \num{96.90}\\
 &  & $\text{C\&W}_{\epsilon=16/255}$ & \num{66.60} & \num{0.70} & \num{75.10} & \num{97.30} & \num{78.30} & \num{87.56} & \num{42.81} & \num{87.90} & \num{89.50}\\
 &  & $\text{C\&W}_{\epsilon=8/255}$  & \num{66.60} & \num{0.60} & \num{75.00} & \num{97.30} & \num{75.90} & \num{88.80} & \num{44.04} & \num{83.90} & \num{89.30}\\
 &  & $\text{AA}_{\epsilon=16/255}$   & \num{66.60} & \num{0.40} & \num{74.60} & \num{98.00} & \num{99.00} & \num{85.94} & \num{13.88} & \num{87.80} & \num{90.50}\\
 &  & $\text{AA}_{\epsilon=8/255}$    & \num{66.60} & \num{0.40} & \num{72.90} & \num{97.30} & \num{90.80} & \num{90.82} & \num{26.06} & \num{86.60} & \num{89.10}\\ \cmidrule{2-12}
 & \multirow[c]{7}{*}{\vspace{-3pt}VGG-16}
 & $\text{PGD}_{\epsilon=16/255}$ & \num{70.00} & \num{45.40} & \num{78.30} & \num{67.60} & \num{99.60} & \num{92.14} & \num{71.66} & \num{93.90} & \num{94.20}\\
 &  & $\text{PGD}_{\epsilon=8/255}$  & \num{82.70} & \num{56.70} & \num{79.80} & \num{70.70} & \num{98.90} & \num{95.01} & \num{74.26} & \num{90.30} & \num{94.40}\\
 &  & $\text{C\&W}_{\epsilon=16/255}$ & \num{11.20} & \num{0.00} & \num{76.70} & \num{67.40} & \num{94.80} & \num{67.33} & \num{45.36} & \num{70.00} & \num{79.80}\\
 &  & $\text{C\&W}_{\epsilon=8/255}$  & \num{47.10} & \num{0.00} & \num{75.10} & \num{68.30} & \num{86.90} & \num{73.40} & \num{45.70} & \num{75.30} & \num{79.90}\\
 &  & $\text{AA}_{\epsilon=16/255}$   & \num{20.20} & \num{0.00} & \num{72.40} & \num{67.40} & \num{98.90} & \num{67.24} & \num{44.40} & \num{70.80} & \num{75.20}\\
 &  & $\text{AA}_{\epsilon=8/255}$    & \num{39.40} & \num{0.00} & \num{72.30} & \num{67.80} & \num{89.80} & \num{74.84} & \num{45.30} & \num{70.10} & \num{81.50}\\ \cmidrule{2-12}
 & \multirow[c]{7}{*}{\vspace{-3pt}ViT}
 & $\text{PGD}_{\epsilon=16/255}$ & \num{98.50} & \num{99.70} & \num{95.80} & \num{66.60} & \num{99.80} & \num{98.46} & \num{98.55} & \num{97.90} & \num{99.40}\\
 &  & $\text{PGD}_{\epsilon=8/255}$  & \num{99.50} & \num{98.90} & \num{94.10} & \num{66.80} & \num{99.80} & \num{97.89} & \num{98.14} & \num{92.80} & \num{99.10}\\
 &  & $\text{C\&W}_{\epsilon=16/255}$ & \num{86.80} & \num{82.90} & \num{83.00} & \num{66.60} & \num{87.70} & \num{89.19} & \num{92.89} & \num{88.40} & \num{93.20}\\
 &  & $\text{C\&W}_{\epsilon=8/255}$  & \num{86.40} & \num{81.70} & \num{83.10} & \num{66.60} & \num{85.50} & \num{90.35} & \num{93.12} & \num{87.00} & \num{92.70}\\
 &  & $\text{AA}_{\epsilon=16/255}$   & \num{84.90} & \num{80.50} & \num{77.60} & \num{66.70} & \num{98.70} & \num{91.45} & \num{91.82} & \num{86.10} & \num{94.00}\\
 &  & $\text{AA}_{\epsilon=8/255}$    & \num{86.90} & \num{80.50} & \num{80.20} & \num{66.40} & \num{91.10} & \num{93.02} & \num{93.51} & \num{86.40} & \num{92.60}\\
\midrule
\multirow[c]{21}{*}{\vspace{-12pt}\textbf{ImageNet}}
 & \multirow[c]{7}{*}{\vspace{-3pt}ResNet-50}
 & $\text{PGD}_{\epsilon=16/255}$ & \num{82.20} & \num{23.40} & \num{92.10} & \num{99.40} & \num{99.70} & \num{94.74} & \num{65.22} & \num{95.20} & \num{96.40}\\
 &  & $\text{PGD}_{\epsilon=8/255}$  & \num{94.40} & \num{30.00} & \num{89.00} & \num{99.40} & \num{92.30} & \num{95.51} & \num{76.82} & \num{93.40} & \num{96.90}\\
 &  & $\text{C\&W}_{\epsilon=16/255}$ & \num{72.40} & \num{2.20} & \num{84.20} & \num{98.80} & \num{66.80} & \num{94.29} & \num{31.67} & \num{92.90} & \num{93.50}\\
 &  & $\text{C\&W}_{\epsilon=8/255}$  & \num{77.10} & \num{2.20} & \num{83.80} & \num{98.90} & \num{66.80} & \num{94.62} & \num{36.12} & \num{93.40} & \num{93.70}\\
 &  & $\text{AA}_{\epsilon=16/255}$   & \num{62.60} & \num{0.00} & \num{79.40} & \num{99.40} & \num{94.90} & \num{93.41} & \num{0.00}  & \num{87.90} & \num{91.50}\\
 &  & $\text{AA}_{\epsilon=8/255}$    & \num{72.90} & \num{0.00} & \num{81.60} & \num{99.40} & \num{70.60} & \num{95.21} & \num{14.00} & \num{88.90} & \num{92.60}\\ \cmidrule{2-12}
 & \multirow[c]{7}{*}{\vspace{-3pt}VGG-16}
 & $\text{PGD}_{\epsilon=16/255}$ & \num{70.30} & \num{24.50} & \num{92.50} & \num{99.10} & \num{99.20} & \num{89.21} & \num{73.17} & \num{91.20} & \num{96.10}\\
 &  & $\text{PGD}_{\epsilon=8/255}$  & \num{84.40} & \num{26.90} & \num{89.50} & \num{98.60} & \num{88.30} & \num{95.26} & \num{79.77} & \num{90.60} & \num{95.60}\\
 &  & $\text{C\&W}_{\epsilon=16/255}$ & \num{2.70} & \num{0.00} & \num{86.50} & \num{99.10} & \num{78.80} & \num{76.25} & \num{14.29} & \num{91.50} & \num{90.40}\\
 &  & $\text{C\&W}_{\epsilon=8/255}$  & \num{13.20} & \num{0.00} & \num{85.60} & \num{98.60} & \num{68.30} & \num{83.19} & \num{11.76} & \num{88.60} & \num{89.90}\\
 &  & $\text{AA}_{\epsilon=16/255}$   & \num{22.00} & \num{2.20} & \num{83.70} & \num{98.60} & \num{94.00} & \num{76.64} & \num{15.93} & \num{88.20} & \num{80.00}\\
 &  & $\text{AA}_{\epsilon=8/255}$    & \num{10.20} & \num{2.20} & \num{81.90} & \num{98.00} & \num{73.10} & \num{86.75} & \num{15.93} & \num{85.70} & \num{81.50}\\ \cmidrule{2-12}
 & \multirow[c]{7}{*}{\vspace{-3pt}ViT}
 & $\text{PGD}_{\epsilon=16/255}$ & \num{99.40} & \num{99.40} & \num{94.50} & \num{66.40} & \num{99.40} & \num{97.79} & \num{99.72} & \num{97.10} & \num{98.80}\\
 &  & $\text{PGD}_{\epsilon=8/255}$  & \num{99.40} & \num{99.40} & \num{94.10} & \num{66.40} & \num{98.00} & \num{98.05} & \num{98.90} & \num{93.80} & \num{98.90}\\
 &  & $\text{C\&W}_{\epsilon=16/255}$ & \num{98.80} & \num{98.90} & \num{91.90} & \num{66.40} & \num{77.30} & \num{96.38} & \num{98.60} & \num{96.30} & \num{96.20}\\
 &  & $\text{C\&W}_{\epsilon=8/255}$  & \num{98.90} & \num{98.90} & \num{91.60} & \num{66.40} & \num{74.60} & \num{96.38} & \num{98.60} & \num{96.00} & \num{95.90}\\
 &  & $\text{AA}_{\epsilon=16/255}$   & \num{98.90} & \num{98.60} & \num{89.30} & \num{66.40} & \num{95.20} & \num{95.80} & \num{97.79} & \num{94.80} & \num{96.20}\\
 &  & $\text{AA}_{\epsilon=8/255}$    & \num{98.90} & \num{98.60} & \num{91.80} & \num{66.40} & \num{78.60} & \num{96.90} & \num{98.88} & \num{95.70} & \num{95.40}\\
\midrule
\multicolumn{3}{c|}{\textbf{Average}} &
\num{74.63} & \num{48.35} & \num{82.09} & \num{78.28} &
\underline{\num{88.91}} & \num{86.40} & \num{65.27} & \num{87.27}{\textcolor{Green}{$\uparrow$\scriptsize\textbf{5.2}}} & \textbf{\num{90.13}}{\textcolor{Green}{$\uparrow$\scriptsize\textbf{11.8}}}\\
\bottomrule
\end{tabular}}%
} 
\end{table*}

{

\sisetup{
  round-mode      = places ,
  round-precision = 2
}

\begin{table*}[ht!]
\centering
\caption{Comparative F1‑scores of the best and most SotA adversarial detectors according to Table~\ref{tab:exp-results}: DNR \citep{sotgiu2020deep}, DKNN \citep{papernot2018deep}, AEAE \citep{liu2022nowhere} and our proposed method \textbf{U‑CAN+DNR} under the \emph{adaptive} setting, using ADA‑DKNN \citep{sitawarin2019robustness}, ADA‑AEAE, ADA‑DCTFF, and ADA‑SD attacks. The \textbf{boldface} entry denotes the best score, and the \underline{underline} entry denotes the second best.}
\label{tab:adaptive-attacks}
\resizebox{0.775\textwidth}{!}{%
\begin{tabular}{c|c|c|c|cc|cc|cc|G{3cm}}
\toprule
\textbf{Dataset} & \textbf{Model} &
\shortstack[c]{DNR\\(ADA‑DKNN)} &
\shortstack[c]{DKNN\\(ADA‑DKNN)} &
\multicolumn{2}{c|}{\shortstack[c]{AEAE\\(ADA‑AEAE)}} &
\multicolumn{2}{c|}{\shortstack[c]{DCTFF\\(ADA‑DCTFF)}} &
\multicolumn{2}{c|}{\shortstack[c]{SD\\(ADA‑SD)}} &
\shortstack[c]{\textbf{U‑CAN+DNR}\\(ADA‑DKNN)}\\
\cmidrule(lr){5-6}\cmidrule(lr){7-8}\cmidrule(lr){9-10}
&&&& \small{$\epsilon=16/255$} & \small{$\epsilon=8/255$}
& \small{$\epsilon=16/255$} & \small{$\epsilon=8/255$}
& \small{$\epsilon=16/255$} & \small{$\epsilon=8/255$} &\\
\midrule
\multirow{3}{*}{CIFAR‑10}
 & ResNet‑50 & \num{69.1} & \num{75.4} & \num{66.7} & \num{66.7} & \num{65.83541147132169} & \num{65.66791510611736} & \num{65.35141800246609} & \num{66.50366748166259} & \num{69.7}\\
 & VGG‑16    & \num{66.7} & \num{74.1} & \num{66.4} & \num{66.4} & \num{61.756373937677054} & \num{62.03966005665723} & \num{66.32390745501286} & \num{66.66666666666666} & \num{73.2}\\
 & ViT       & \num{67.8} & \num{72.9} & \num{66.5} & \num{66.5} & \num{65.58197747183979} & \num{66.74937965260547} & \num{81.49350649350649} & \num{82.44766505636072} & \num{80.6}\\
\midrule
\multirow{3}{*}{Mammals}
 & ResNet‑50 & \num{67.5} & \num{67.3} & \num{67.7} & \num{67.4} & \num{66.24605678233438} & \num{65.99947326836977} & \num{43.20927984533591} & \num{44.71038774533269} & \num{66.6}\\
 & VGG‑16    & \num{66.6} & \num{68.7} & \num{66.6} & \num{66.6} & \num{66.43037974683544} & \num{66.43037974683544} & \num{45.0200656694637} & \num{45.52727272727273} & \num{66.6}\\
 & ViT       & \num{66.7} & \num{68.2} & \num{66.7} & \num{66.7} & \num{66.7337188835806} & \num{66.59954694185754} & \num{93.08452250274424} & \num{93.16270566727603} & \num{66.6}\\
\midrule
\multirow{3}{*}{ImageNet}
 & ResNet‑50 & \num{67.2} & \num{70.8} & \num{69.0} & \num{66.4} & \num{30.303030303030297} & \num{36.734693877551024} & \num{25.352112676056336} & \num{34.666666666666664} & \num{66.4}\\
 & VGG‑16    & \num{66.8} & \num{72.3} & \num{66.3} & \num{66.4} & \num{1.4981273408239701} & \num{1.4981273408239701} & \num{15.929203539823} & \num{11.76470588235294} & \num{67.8}\\
 & ViT       & \num{68.8} & \num{70.9} & \num{66.4} & \num{66.4} & \num{69.21241050119332} & \num{70.64676616915422} & \num{98.59943977591035} & \num{98.59943977591035} & \num{72.2}\\
\midrule
\multicolumn{2}{c|}{\textbf{Average}} &
\num{67.5} & \textbf{\num{71.2}} & \num{66.9} & \num{66.6} &
\num{54.84416515984851} & \num{55.81843801777467} &
\num{59.373717328924336} & \num{60.44990862994459} &
\underline{\num{70.0}}\\
\bottomrule
\end{tabular}}
\end{table*}
}

\paragraph{Comparative evaluation under adaptive attacks.}
We assessed the six newest and strongest detectors: DKNN, DNR, AEAE, DCTFF, SD, and our top variant U-CAN+DNR from Table \ref{tab:exp-results}. As before, PR curves were generated for every setting, and the F1-maximizing threshold was chosen. Three adaptive attacks were employed: ADA-DKNN for feature-space methods (DKNN, DNR, U-CAN+DNR), ADA-AEAE for AEAE, ADA-DCTFF for DCTFF, and ADA-SD for SD (see Section \ref{adaptive_attacks_definition}). Table \ref{tab:adaptive-attacks} shows a similar performance drop across detectors: averaged over all datasets and architectures, DKNN reaches 71.2\% F1, U-CAN+DNR 70.0\%, and the remaining baselines produce less than 67\% F1 scores.

Notably, the ability to refine intermediate features in U-CAN integrates seamlessly with other feature-based detectors such as DNR \citep{sotgiu2020deep}, without incurring an additional robustness penalty under the scenario of adaptive attacks.

\section{Conclusions}
\label{sec:conclusions}
We introduced \emph{U-CAN}, an unsupervised adversarial detection framework that attaches ArcFace-based auxiliary blocks to the intermediate layers of a frozen target model. Trained solely on benign data, U-CAN requires no backbone retraining or adversarial examples. Experiments on three datasets: \citep{krizhevsky2009learning,asaniczka_2023}, and a subset of \citep{russakovsky2015imagenet}, three architectures \citep{he2016deep,simonyan2014very,dosovitskiy2020image}, three powerful, well-known attacks \citep{madry2017towards,carlini2017towards,croce2020reliable}, show that projecting intermediate features into a contrastive, less noisy space substantially boosts detection performance by making small perturbations easier to spot, while incurring only negligible additional latency. When paired with off-the-shelf layer-wise detectors, U-CAN consistently achieves higher F1 scores and greater stability than the original baselines. These results suggest that U-CAN can significantly enhance adversarial detection in safety-critical applications.

\section{Future Work and Limitations}
\label{sec:future_work}
There are several promising extensions for U-CAN.  
\textbf{(i)} Design alternative aggregation detectors—specifically, ones that exploit the temporal relationship between the layer-wise features, may further improve performance. \textbf{(ii)} Validate U-CAN on additional tasks and modalities—e.g., object recognition, object detection, and audio-, text-, or image-based generative models. \textbf{(iii)} Study attacks targeting the nearest label and test whether U-CAN still distinguishes subtle intra-class or subclass differences. In summary, our method lays a generic and solid foundation for adversarial detection but may be strengthened through complementary techniques and a deeper analysis of feature interactions.

\newpage
\bibliographystyle{ieeenat_fullname}
\bibliography{refs}

\begin{thebibliography}{74}
\providecommand{\natexlab}[1]{#1}
\providecommand{\url}[1]{\texttt{#1}}
\expandafter\ifx\csname urlstyle\endcsname\relax
  \providecommand{\doi}[1]{doi: #1}\else
  \providecommand{\doi}{doi: \begingroup \urlstyle{rm}\Url}\fi

\bibitem[Alter et~al.(2025)Alter, Lapid, and Sipper]{alter2025on}
Tal Alter, Raz Lapid, and Moshe Sipper.
\newblock On the robustness of kolmogorov-arnold networks: An adversarial perspective.
\newblock \emph{Transactions on Machine Learning Research}, 2025.

\bibitem[Andriushchenko and Flammarion(2020)]{andriushchenko2020understanding}
Maksym Andriushchenko and Nicolas Flammarion.
\newblock Understanding and improving fast adversarial training.
\newblock \emph{Advances in Neural Information Processing Systems}, 33:\penalty0 16048--16059, 2020.

\bibitem[Andriushchenko et~al.(2020)Andriushchenko, Croce, Flammarion, and Hein]{andriushchenko2020square}
Maksym Andriushchenko, Francesco Croce, Nicolas Flammarion, and Matthias Hein.
\newblock Square attack: a query-efficient black-box adversarial attack via random search.
\newblock In \emph{European conference on computer vision}, pages 484--501. Springer, 2020.

\bibitem[Asaniczka(2023)]{asaniczka_2023}
Asaniczka.
\newblock Mammals image classification dataset (45 animals), 2023.

\bibitem[Bachute and Subhedar(2021)]{bachute2021autonomous}
Mrinal~R Bachute and Javed~M Subhedar.
\newblock Autonomous driving architectures: insights of machine learning and deep learning algorithms.
\newblock \emph{Machine Learning with Applications}, 6:\penalty0 100164, 2021.

\bibitem[Botteghi et~al.(2025)Botteghi, Poel, and Brune]{botteghi2025unsupervised}
Nicol{\`o} Botteghi, Mannes Poel, and Christoph Brune.
\newblock Unsupervised representation learning in deep reinforcement learning: A review.
\newblock \emph{IEEE Control Systems}, 45\penalty0 (2):\penalty0 26--68, 2025.

\bibitem[Carlini and Wagner(2017{\natexlab{a}})]{carlini2017adversarial}
Nicholas Carlini and David Wagner.
\newblock Adversarial examples are not easily detected: Bypassing ten detection methods.
\newblock In \emph{Proceedings of the 10th ACM workshop on artificial intelligence and security}, pages 3--14, 2017{\natexlab{a}}.

\bibitem[Carlini and Wagner(2017{\natexlab{b}})]{carlini2017towards}
Nicholas Carlini and David Wagner.
\newblock Towards evaluating the robustness of neural networks.
\newblock In \emph{2017 ieee symposium on security and privacy (sp)}, pages 39--57. Ieee, 2017{\natexlab{b}}.

\bibitem[Carrara et~al.(2017)Carrara, Falchi, Caldelli, Amato, Fumarola, and Becarelli]{carrara2017detecting}
Fabio Carrara, Fabrizio Falchi, Roberto Caldelli, Giuseppe Amato, Roberta Fumarola, and Rudy Becarelli.
\newblock Detecting adversarial example attacks to deep neural networks.
\newblock In \emph{Proceedings of the 15th international workshop on content-based multimedia indexing}, pages 1--7, 2017.

\bibitem[Carrara et~al.(2018)Carrara, Becarelli, Caldelli, Falchi, and Amato]{carrara2018adversarial}
Fabio Carrara, Rudy Becarelli, Roberto Caldelli, Fabrizio Falchi, and Giuseppe Amato.
\newblock Adversarial examples detection in features distance spaces.
\newblock In \emph{Proceedings of the European conference on computer vision (ECCV) workshops}, pages 0--0, 2018.

\bibitem[Chen et~al.(2020)Chen, Kornblith, Norouzi, and Hinton]{chen2020simple}
Ting Chen, Simon Kornblith, Mohammad Norouzi, and Geoffrey Hinton.
\newblock A simple framework for contrastive learning of visual representations.
\newblock In \emph{International conference on machine learning}, pages 1597--1607. PMLR, 2020.

\bibitem[Chiang et~al.(2020)Chiang, Ni, Abdelkader, Zhu, Studor, and Goldstein]{chiangcertified}
Ping-yeh Chiang, Renkun Ni, Ahmed Abdelkader, Chen Zhu, Christoph Studor, and Tom Goldstein.
\newblock Certified defenses for adversarial patches.
\newblock In \emph{International Conference on Learning Representations}, 2020.

\bibitem[Chitic et~al.(2023)Chitic, Topal, and Leprévost]{app13064068}
Raluca Chitic, Ali~Osman Topal, and Franck Leprévost.
\newblock Shuffledetect: Detecting adversarial images against convolutional neural networks.
\newblock \emph{Applied Sciences}, 13\penalty0 (6), 2023.

\bibitem[Chyou et~al.(2023)Chyou, Su, and Hsu]{chyou2023unsupervised}
Chien~Cheng Chyou, Hung-Ting Su, and Winston~H Hsu.
\newblock Unsupervised adversarial detection without extra model: training loss should change.
\newblock \emph{arXiv preprint arXiv:2308.03243}, 2023.

\bibitem[Cohen et~al.(2019)Cohen, Rosenfeld, and Kolter]{cohen2019certified}
Jeremy Cohen, Elan Rosenfeld, and Zico Kolter.
\newblock Certified adversarial robustness via randomized smoothing.
\newblock In \emph{international conference on machine learning}, pages 1310--1320. PMLR, 2019.

\bibitem[Craighero et~al.(2023)Craighero, Angaroni, Stella, Damiani, Antoniotti, and Graudenzi]{craighero2023unity}
Francesco Craighero, Fabrizio Angaroni, Fabio Stella, Chiara Damiani, Marco Antoniotti, and Alex Graudenzi.
\newblock Unity is strength: Improving the detection of adversarial examples with ensemble approaches.
\newblock \emph{Neurocomputing}, 554:\penalty0 126576, 2023.

\bibitem[Croce and Hein(2020{\natexlab{a}})]{croce2020minimally}
Francesco Croce and Matthias Hein.
\newblock Minimally distorted adversarial examples with a fast adaptive boundary attack.
\newblock In \emph{International Conference on Machine Learning}, pages 2196--2205. PMLR, 2020{\natexlab{a}}.

\bibitem[Croce and Hein(2020{\natexlab{b}})]{croce2020reliable}
Francesco Croce and Matthias Hein.
\newblock Reliable evaluation of adversarial robustness with an ensemble of diverse parameter-free attacks.
\newblock In \emph{International conference on machine learning}, pages 2206--2216. PMLR, 2020{\natexlab{b}}.

\bibitem[Dathathri et~al.(2018)Dathathri, Zheng, Yin, Murray, and Yue]{dathathri2018detecting}
Sumanth Dathathri, Stephan Zheng, Tianwei Yin, Richard~M Murray, and Yisong Yue.
\newblock Detecting adversarial examples via neural fingerprinting.
\newblock \emph{arXiv preprint arXiv:1803.03870}, 2018.

\bibitem[Deng et~al.(2022)Deng, Guo, Yang, Xue, Kotsia, and Zafeiriou]{Deng_2022}
Jiankang Deng, Jia Guo, Jing Yang, Niannan Xue, Irene Kotsia, and Stefanos Zafeiriou.
\newblock Arcface: Additive angular margin loss for deep face recognition.
\newblock \emph{IEEE Transactions on Pattern Analysis and Machine Intelligence}, 44\penalty0 (10):\penalty0 5962–5979, 2022.

\bibitem[Dosovitskiy(2020)]{dosovitskiy2020image}
Alexey Dosovitskiy.
\newblock An image is worth 16x16 words: Transformers for image recognition at scale.
\newblock \emph{arXiv preprint arXiv:2010.11929}, 2020.

\bibitem[Esteva et~al.(2019)Esteva, Robicquet, Ramsundar, Kuleshov, DePristo, Chou, Cui, Corrado, Thrun, and Dean]{esteva2019guide}
Andre Esteva, Alexandre Robicquet, Bharath Ramsundar, Volodymyr Kuleshov, Mark DePristo, Katherine Chou, Claire Cui, Greg Corrado, Sebastian Thrun, and Jeff Dean.
\newblock A guide to deep learning in healthcare.
\newblock \emph{Nature medicine}, 25\penalty0 (1):\penalty0 24--29, 2019.

\bibitem[Fan et~al.(2025)Fan, Tong, Zhu, Sinha, Liu, Chen, Rabbat, Ballas, LeCun, Bar, et~al.]{fan2025scaling}
David Fan, Shengbang Tong, Jiachen Zhu, Koustuv Sinha, Zhuang Liu, Xinlei Chen, Michael Rabbat, Nicolas Ballas, Yann LeCun, Amir Bar, et~al.
\newblock Scaling language-free visual representation learning.
\newblock \emph{arXiv preprint arXiv:2504.01017}, 2025.

\bibitem[Feinman et~al.(2017)Feinman, Curtin, Shintre, and Gardner]{feinman2017detecting}
Reuben Feinman, Ryan~R Curtin, Saurabh Shintre, and Andrew~B Gardner.
\newblock Detecting adversarial samples from artifacts.
\newblock \emph{arXiv preprint arXiv:1703.00410}, 2017.

\bibitem[Finlayson et~al.(2019)Finlayson, Bowers, Ito, Zittrain, Beam, and Kohane]{finlayson2019adversarial}
Samuel~G Finlayson, John~D Bowers, Joichi Ito, Jonathan~L Zittrain, Andrew~L Beam, and Isaac~S Kohane.
\newblock Adversarial attacks on medical machine learning.
\newblock \emph{Science}, 363\penalty0 (6433):\penalty0 1287--1289, 2019.

\bibitem[Ford and Siraj(2014)]{ford2014applications}
Vitaly Ford and Ambareen Siraj.
\newblock Applications of machine learning in cyber security.
\newblock In \emph{Proceedings of the 27th international conference on computer applications in industry and engineering}. IEEE Xplore Kota Kinabalu, Malaysia, 2014.

\bibitem[Goodfellow et~al.(2014)Goodfellow, Shlens, and Szegedy]{goodfellow2014explaining}
Ian~J Goodfellow, Jonathon Shlens, and Christian Szegedy.
\newblock Explaining and harnessing adversarial examples.
\newblock \emph{arXiv preprint arXiv:1412.6572}, 2014.

\bibitem[Graves and Graves(2012)]{graves2012long}
Alex Graves and Alex Graves.
\newblock Long short-term memory.
\newblock \emph{Supervised sequence labelling with recurrent neural networks}, pages 37--45, 2012.

\bibitem[Grigorescu et~al.(2020)Grigorescu, Trasnea, Cocias, and Macesanu]{grigorescu2020survey}
Sorin Grigorescu, Bogdan Trasnea, Tiberiu Cocias, and Gigel Macesanu.
\newblock A survey of deep learning techniques for autonomous driving.
\newblock \emph{Journal of field robotics}, 37\penalty0 (3):\penalty0 362--386, 2020.

\bibitem[Grosse et~al.(2017)Grosse, Manoharan, Papernot, Backes, and McDaniel]{grosse2017statistical}
Kathrin Grosse, Praveen Manoharan, Nicolas Papernot, Michael Backes, and Patrick McDaniel.
\newblock On the (statistical) detection of adversarial examples.
\newblock \emph{arXiv preprint arXiv:1702.06280}, 2017.

\bibitem[Guo et~al.(2019)Guo, Gardner, You, Wilson, and Weinberger]{guo2019simple}
Chuan Guo, Jacob Gardner, Yurong You, Andrew~Gordon Wilson, and Kilian Weinberger.
\newblock Simple black-box adversarial attacks.
\newblock In \emph{International conference on machine learning}, pages 2484--2493. PMLR, 2019.

\bibitem[He et~al.(2016)He, Zhang, Ren, and Sun]{he2016deep}
Kaiming He, Xiangyu Zhang, Shaoqing Ren, and Jian Sun.
\newblock Deep residual learning for image recognition.
\newblock In \emph{Proceedings of the IEEE conference on computer vision and pattern recognition}, pages 770--778, 2016.

\bibitem[Hendrycks and Gimpel(2016)]{hendrycks2016baseline}
Dan Hendrycks and Kevin Gimpel.
\newblock A baseline for detecting misclassified and out-of-distribution examples in neural networks.
\newblock \emph{arXiv preprint arXiv:1610.02136}, 2016.

\bibitem[Kazoom et~al.(2025)Kazoom, Lapid, Sipper, and Hadar]{kazoom2025dont}
Roie Kazoom, Raz Lapid, Moshe Sipper, and Ofer Hadar.
\newblock Don{\textquoteright}t lag, {RAG}: Training-free adversarial detection using {RAG}.
\newblock In \emph{The 1st Workshop on Vector Databases}, 2025.

\bibitem[Khosla et~al.(2020)Khosla, Teterwak, Wang, Sarna, Tian, Isola, Maschinot, Liu, and Krishnan]{khosla2020supervised}
Prannay Khosla, Piotr Teterwak, Chen Wang, Aaron Sarna, Yonglong Tian, Phillip Isola, Aaron Maschinot, Ce Liu, and Dilip Krishnan.
\newblock Supervised contrastive learning.
\newblock \emph{Advances in neural information processing systems}, 33:\penalty0 18661--18673, 2020.

\bibitem[Krizhevsky et~al.(2009)Krizhevsky, Hinton, et~al.]{krizhevsky2009learning}
Alex Krizhevsky, Geoffrey Hinton, et~al.
\newblock Learning multiple layers of features from tiny images.
\newblock Technical report, University of Toronto, Toronto, ON, Canada, 2009.
\newblock Available at: \url{https://www.cs.toronto.edu/~kriz/learning-features-2009-TR.pdf}.

\bibitem[Lapid and Sipper(2023)]{lapid2023see}
Raz Lapid and Moshe Sipper.
\newblock I see dead people: Gray-box adversarial attack on image-to-text models.
\newblock In \emph{Joint European Conference on Machine Learning and Knowledge Discovery in Databases}, pages 277--289. Springer, 2023.

\bibitem[Lapid and Sipper(2024)]{lapid2023patch}
Raz Lapid and Moshe Sipper.
\newblock Patch of invisibility: Naturalistic black-box adversarial attacks on object detectors.
\newblock In \emph{6th Workshop on Machine Learning for Cybersecurity, part of ECMLPKDD 2024}, 2024.

\bibitem[Lapid et~al.(2022)Lapid, Haramaty, and Sipper]{lapid2022evolutionary}
Raz Lapid, Zvika Haramaty, and Moshe Sipper.
\newblock An evolutionary, gradient-free, query-efficient, black-box algorithm for generating adversarial instances in deep convolutional neural networks.
\newblock \emph{Algorithms}, 15\penalty0 (11):\penalty0 407, 2022.

\bibitem[Lapid et~al.(2024{\natexlab{a}})Lapid, Dubin, and Sipper]{lapid2024fortify}
Raz Lapid, Almog Dubin, and Moshe Sipper.
\newblock Fortify the guardian, not the treasure: Resilient adversarial detectors.
\newblock \emph{Mathematics}, 12\penalty0 (22):\penalty0 3451, 2024{\natexlab{a}}.

\bibitem[Lapid et~al.(2024{\natexlab{b}})Lapid, Langberg, and Sipper]{lapid2024open}
Raz Lapid, Ron Langberg, and Moshe Sipper.
\newblock Open sesame! universal black-box jailbreaking of large language models.
\newblock In \emph{ICLR 2024 Workshop on Secure and Trustworthy Large Language Models}, 2024{\natexlab{b}}.

\bibitem[Lee et~al.(2018)Lee, Lee, Lee, and Shin]{lee2018simple}
Kimin Lee, Kibok Lee, Honglak Lee, and Jinwoo Shin.
\newblock A simple unified framework for detecting out-of-distribution samples and adversarial attacks.
\newblock \emph{Advances in neural information processing systems}, 31, 2018.

\bibitem[Li et~al.(2023)Li, Xie, and Li]{li2023sok}
Linyi Li, Tao Xie, and Bo Li.
\newblock Sok: Certified robustness for deep neural networks.
\newblock In \emph{2023 IEEE symposium on security and privacy (SP)}, pages 1289--1310. IEEE, 2023.

\bibitem[Li and Li(2017)]{li2017adversarial}
Xin Li and Fuxin Li.
\newblock Adversarial examples detection in deep networks with convolutional filter statistics.
\newblock In \emph{Proceedings of the IEEE international conference on computer vision}, pages 5764--5772, 2017.

\bibitem[Lin(2002)]{lin2002divergence}
Jianhua Lin.
\newblock Divergence measures based on the shannon entropy.
\newblock \emph{IEEE Transactions on Information theory}, 37\penalty0 (1):\penalty0 145--151, 2002.

\bibitem[Liu et~al.(2008)Liu, Ting, and Zhou]{liu2008isolation}
Fei~Tony Liu, Kai~Ming Ting, and Zhi-Hua Zhou.
\newblock Isolation forest.
\newblock In \emph{2008 eighth ieee international conference on data mining}, pages 413--422. IEEE, 2008.

\bibitem[Liu et~al.(2022{\natexlab{a}})Liu, Zhao, Ji, Peng, Guo, and Liu]{liu2022feature}
Hui Liu, Bo Zhao, Minzhi Ji, Yuefeng Peng, Jiabao Guo, and Peng Liu.
\newblock Feature-filter: Detecting adversarial examples by filtering out recessive features.
\newblock \emph{Applied Soft Computing}, 124:\penalty0 109027, 2022{\natexlab{a}}.

\bibitem[Liu et~al.(2022{\natexlab{b}})Liu, Zhao, Zhang, and Liu]{liu2022nowhere}
Hui Liu, Bo Zhao, Kehuan Zhang, and Peng Liu.
\newblock Nowhere to hide: A lightweight unsupervised detector against adversarial examples.
\newblock \emph{arXiv preprint arXiv:2210.08579}, 2022{\natexlab{b}}.

\bibitem[Ma et~al.(2018)Ma, Li, Wang, Erfani, Wijewickrema, Schoenebeck, Song, Houle, and Bailey]{ma2018characterizing}
Xingjun Ma, Bo Li, Yisen Wang, Sarah~M Erfani, Sudanthi Wijewickrema, Grant Schoenebeck, Dawn Song, Michael~E Houle, and James Bailey.
\newblock Characterizing adversarial subspaces using local intrinsic dimensionality.
\newblock \emph{arXiv preprint arXiv:1801.02613}, 2018.

\bibitem[Madry(2017)]{madry2017towards}
Aleksander Madry.
\newblock Towards deep learning models resistant to adversarial attacks.
\newblock \emph{arXiv preprint arXiv:1706.06083}, 2017.

\bibitem[Meng and Chen(2017)]{meng2017magnet}
Dongyu Meng and Hao Chen.
\newblock Magnet: a two-pronged defense against adversarial examples.
\newblock In \emph{Proceedings of the 2017 ACM SIGSAC conference on computer and communications security}, pages 135--147, 2017.

\bibitem[Metzen et~al.(2022)Metzen, Genewein, Fischer, and Bischoff]{metzen2022detecting}
Jan~Hendrik Metzen, Tim Genewein, Volker Fischer, and Bastian Bischoff.
\newblock On detecting adversarial perturbations.
\newblock In \emph{International Conference on Learning Representations}, 2022.

\bibitem[Mu et~al.(2025)Mu, Li, Peng, Wang, and Liang]{mu2025robust}
Hua Mu, Chenggang Li, Anjie Peng, Yangyang Wang, and Zhenyu Liang.
\newblock Robust adversarial example detection algorithm based on high-level feature differences.
\newblock \emph{Sensors}, 25\penalty0 (6):\penalty0 1770, 2025.

\bibitem[Oord et~al.(2018)Oord, Li, and Vinyals]{oord2018representation}
Aaron van~den Oord, Yazhe Li, and Oriol Vinyals.
\newblock Representation learning with contrastive predictive coding.
\newblock \emph{arXiv preprint arXiv:1807.03748}, 2018.

\bibitem[Pang et~al.(2022)Pang, Zhang, He, Dong, Su, Chen, Zhu, and Liu]{pang2022two}
Tianyu Pang, Huishuai Zhang, Di He, Yinpeng Dong, Hang Su, Wei Chen, Jun Zhu, and Tie-Yan Liu.
\newblock Two coupled rejection metrics can tell adversarial examples apart.
\newblock In \emph{Proceedings of the IEEE/CVF conference on computer vision and pattern recognition}, pages 15223--15233, 2022.

\bibitem[Papernot and McDaniel(2018)]{papernot2018deep}
Nicolas Papernot and Patrick McDaniel.
\newblock Deep k-nearest neighbors: Towards confident, interpretable and robust deep learning.
\newblock \emph{arXiv preprint arXiv:1803.04765}, 2018.

\bibitem[Pinhasov et~al.(2024)Pinhasov, Lapid, Ohayon, Sipper, and Aperstein]{pinhasov2024xaibased}
Ben Pinhasov, Raz Lapid, Rony Ohayon, Moshe Sipper, and Yehudit Aperstein.
\newblock {XAI}-based detection of adversarial attacks on deepfake detectors.
\newblock \emph{Transactions on Machine Learning Research}, 2024.

\bibitem[Raghunathan et~al.(2018)Raghunathan, Steinhardt, and Liang]{raghunathan2018certified}
Aditi Raghunathan, Jacob Steinhardt, and Percy Liang.
\newblock Certified defenses against adversarial examples.
\newblock In \emph{International Conference on Learning Representations}, 2018.

\bibitem[Roth et~al.(2019)Roth, Kilcher, and Hofmann]{roth2019odds}
Kevin Roth, Yannic Kilcher, and Thomas Hofmann.
\newblock The odds are odd: A statistical test for detecting adversarial examples.
\newblock In \emph{International Conference on Machine Learning}, pages 5498--5507. PMLR, 2019.

\bibitem[Russakovsky et~al.(2015)Russakovsky, Deng, Su, Krause, Satheesh, Ma, Huang, Karpathy, Khosla, Bernstein, et~al.]{russakovsky2015imagenet}
Olga Russakovsky, Jia Deng, Hao Su, Jonathan Krause, Sanjeev Satheesh, Sean Ma, Zhiheng Huang, Andrej Karpathy, Aditya Khosla, Michael Bernstein, et~al.
\newblock Imagenet large scale visual recognition challenge.
\newblock \emph{International journal of computer vision}, 115:\penalty0 211--252, 2015.

\bibitem[Schroff et~al.(2015)Schroff, Kalenichenko, and Philbin]{schroff2015facenet}
Florian Schroff, Dmitry Kalenichenko, and James Philbin.
\newblock Facenet: A unified embedding for face recognition and clustering.
\newblock In \emph{Proceedings of the IEEE conference on computer vision and pattern recognition}, pages 815--823, 2015.

\bibitem[Shaukat et~al.(2020)Shaukat, Luo, Varadharajan, Hameed, and Xu]{shaukat2020survey}
Kamran Shaukat, Suhuai Luo, Vijay Varadharajan, Ibrahim~A Hameed, and Min Xu.
\newblock A survey on machine learning techniques for cyber security in the last decade.
\newblock \emph{IEEE access}, 8:\penalty0 222310--222354, 2020.

\bibitem[Simonyan(2014)]{simonyan2014very}
Karen Simonyan.
\newblock Very deep convolutional networks for large-scale image recognition.
\newblock \emph{arXiv preprint arXiv:1409.1556}, 2014.

\bibitem[Sitawarin and Wagner(2019)]{sitawarin2019robustness}
Chawin Sitawarin and David Wagner.
\newblock On the robustness of deep k-nearest neighbors.
\newblock In \emph{2019 IEEE Security and Privacy Workshops (SPW)}, pages 1--7. IEEE, 2019.

\bibitem[Sotgiu et~al.(2020)Sotgiu, Demontis, Melis, Biggio, Fumera, Feng, and Roli]{sotgiu2020deep}
Angelo Sotgiu, Ambra Demontis, Marco Melis, Battista Biggio, Giorgio Fumera, Xiaoyi Feng, and Fabio Roli.
\newblock Deep neural rejection against adversarial examples.
\newblock \emph{EURASIP Journal on Information Security}, 2020:\penalty0 1--10, 2020.

\bibitem[Su et~al.(2019)Su, Vargas, and Sakurai]{su2019one}
Jiawei Su, Danilo~Vasconcellos Vargas, and Kouichi Sakurai.
\newblock One pixel attack for fooling deep neural networks.
\newblock \emph{IEEE Transactions on Evolutionary Computation}, 23\penalty0 (5):\penalty0 828--841, 2019.

\bibitem[Sun et~al.(2024)Sun, Nwodo, Sugrim, Stavrou, and Wang]{sun2024vitguard}
Shihua Sun, Kenechukwu Nwodo, Shridatt Sugrim, Angelos Stavrou, and Haining Wang.
\newblock Vitguard: Attention-aware detection against adversarial examples for vision transformer.
\newblock \emph{arXiv preprint arXiv:2409.13828}, 2024.

\bibitem[Suzuki(2017)]{suzuki2017overview}
Kenji Suzuki.
\newblock Overview of deep learning in medical imaging.
\newblock \emph{Radiological physics and technology}, 10\penalty0 (3):\penalty0 257--273, 2017.

\bibitem[Szegedy et~al.(2013)Szegedy, Zaremba, Sutskever, Bruna, Erhan, Goodfellow, and Fergus]{szegedy2013intriguing}
Christian Szegedy, Wojciech Zaremba, Ilya Sutskever, Joan Bruna, Dumitru Erhan, Ian Goodfellow, and Rob Fergus.
\newblock Intriguing properties of neural networks.
\newblock \emph{arXiv preprint arXiv:1312.6199}, 2013.

\bibitem[Tamam et~al.(2023)Tamam, Lapid, and Sipper]{vitracktamam2023foiling}
Snir~Vitrack Tamam, Raz Lapid, and Moshe Sipper.
\newblock Foiling explanations in deep neural networks.
\newblock \emph{Transactions on Machine Learning Research}, 2023.

\bibitem[Wang et~al.(2023)Wang, Pang, Du, Lin, Liu, and Yan]{wang2023better}
Zekai Wang, Tianyu Pang, Chao Du, Min Lin, Weiwei Liu, and Shuicheng Yan.
\newblock Better diffusion models further improve adversarial training.
\newblock In \emph{International conference on machine learning}, pages 36246--36263. PMLR, 2023.

\bibitem[Xie et~al.(2019)Xie, Wu, Maaten, Yuille, and He]{xie2019feature}
Cihang Xie, Yuxin Wu, Laurens van~der Maaten, Alan~L Yuille, and Kaiming He.
\newblock Feature denoising for improving adversarial robustness.
\newblock In \emph{Proceedings of the IEEE/CVF conference on computer vision and pattern recognition}, pages 501--509, 2019.

\bibitem[Xu(2017)]{xu2017feature}
W Xu.
\newblock Feature squeezing: Detecting adversarial exa mples in deep neural networks.
\newblock \emph{arXiv preprint arXiv:1704.01155}, 2017.

\bibitem[Yeo et~al.(2021)Yeo, Kar, and Zamir]{yeo2021robustness}
Teresa Yeo, O{\u{g}}uzhan~Fatih Kar, and Amir Zamir.
\newblock Robustness via cross-domain ensembles.
\newblock In \emph{Proceedings of the IEEE/CVF International Conference on Computer Vision}, pages 12189--12199, 2021.

\end{thebibliography}
\end{document}